\documentclass{IEEEtran}
\usepackage{cite}
\usepackage{amsmath,amssymb,amsfonts}
\usepackage{algorithmic}
\usepackage{graphicx}
\usepackage{textcomp}
\usepackage{float}
\usepackage{hyperref}
\usepackage{multirow}
\usepackage{caption}
\usepackage{booktabs}
\usepackage{array}
\usepackage[table]{xcolor}

\usepackage{pict2e}

\newsavebox{\ORCIDlogo}
\savebox{\ORCIDlogo}{%
\setlength{\unitlength}{\dimexpr 1em/256\relax}%
\begin{picture}(256,256)%
  \color[HTML]{A6CE39}\put(128,128){\circle*{256}}%
  \color{white}%
  \put(78.6,199.2){\circle*{20}}%
  \moveto(70.9,176,9)\lineto(86.3,176,9)\lineto(86.3,69.8)\lineto(70.9,69.8)%
  \closepath\fillpath%
  \moveto(108.9,176.9)\lineto(150.5,176.9)%
  \curveto(190.1,176.9)(207.5,148.6)(207.5 ,123.3)%
  \curveto(207.5,95,8)(186,69.7)(150.7,69.7)%
  \lineto(108.9,69.7)%
  \closepath\fillpath%
  \color[HTML]{A6CE39}%
  \moveto(124.3,83.6)\lineto(148.8,83.6)%
  \curveto(183.7,83.6)(191.7,110.1)(191.7,123.3)%
  \curveto(191.7,144.8)(178,163)(148,163)%
  \lineto(124.3,163)%
  \closepath\fillpath%
\end{picture}%
}

\begin{document}

\newcommand{\etal}{\textit{et al}.}
\newcommand\orcidicon[1]{\href{https://orcid.org/#1}{\usebox{\ORCIDlogo}}}

\title{A Review of Deep Learning Approaches for Non-Invasive Cognitive Impairment Detection}

\author{\uppercase{Muath Alsuhaibani\orcidicon{0009-0007-5164-012X},
Ali Pourramezan Fard\orcidicon{0000-0002-3807-0798},
Jian Sun\orcidicon{0000-0002-9367-0892},
Farida Far Poor\orcidicon{0009-0009-8290-0515},
Peter S. Pressman,
and Mohammad H. Mahoor\orcidicon{0000-0001-8923-4660}}.
\thanks{M. Alsuhaibani, A.P. Fard, J. Sun, F. Far Poor, and M.H. Mahoor are with the Ritchie School of Engineering and Computer Science, University of Denver, Denver, CO 80210, USA}
\thanks{P.S. Pressman is with the Department of Neurology, Behavioral Neurology Section, University of Colorado Anschutz Medical Center, Aurora, CO 80045, USA}}

\maketitle

\begin{abstract}
This review paper explores recent advances in deep learning approaches for non-invasive cognitive impairment detection. We examine various non-invasive indicators of cognitive decline, including speech and language, facial, and motoric mobility. The paper provides an overview of relevant datasets, feature-extracting techniques, and deep-learning architectures applied to this domain. We have analyzed the performance of different methods across modalities and observed that speech and language-based methods generally achieved the highest detection performance. Studies combining acoustic and linguistic features tended to outperform those using a single modality. Facial analysis methods showed promise for visual modalities but were less extensively studied. Most papers focused on binary classification (impaired vs. non-impaired), with fewer addressing multi-class or regression tasks. Transfer learning and pre-trained language models emerged as popular and effective techniques, especially for linguistic analysis. Despite significant progress, several challenges remain, including data standardization and accessibility, model explainability, longitudinal analysis limitations, and clinical adaptation. Lastly, we propose future research directions, such as investigating language-agnostic speech analysis methods, developing multi-modal diagnostic systems, and addressing ethical considerations in AI-assisted healthcare. By synthesizing current trends and identifying key obstacles, this review aims to guide further development of deep learning-based cognitive impairment detection systems to improve early diagnosis and ultimately patient outcomes.
\end{abstract}



\section{Introduction}\label{sec.Introduction}

Cognitive impairment poses significant challenges for individuals, families, and healthcare systems worldwide \cite{better2024alzheimer}. As the population of older adults in the United States grows, the prevalence of cognitive impairment related to Alzheimer's disease (AD), AD-related dementia (ADRD), and mild cognitive impairment (MCI), which often progresses to AD/ADRD, is expected to rise, necessitating cost-effective screening tools and early treatment strategies. Current diagnostic methods include clinical evaluations, neuropsychological assessments, and advanced neuroimaging techniques such as Magnetic Resonance Imaging (MRI), Positron Emission Tomography (PET), and Computed Tomography (CT) scans \cite{pressman2024alzheimer}. While effective, these diagnostic methods are expensive and require specialized personnel. Alternatively, scientists have explored cost-effective methods, particularly those leveraging innovative technologies powered by artificial intelligence (AI) and machine learning (ML). 

In recent years, deep learning models have made remarkable progress in diverse domains, including Computer Vision (CV) and Natural Language Processing (NLP), with various implications for healthcare \cite{shamshirband2021review}. Unlike traditional machine learning approaches that separate feature extraction and model learning, deep learning builds end-to-end systems in which neural networks concurrently learn and extract discriminant features directly from the data during the training process. Deep learning models are enhancing various domains, including the detection of pathological indicators among clinically diagnosed patients~\cite{haug2023artificial}, and may enhance detection in early disease stages~\cite{nevler2024changes, petti2023generalizability}. Furthermore, machine learning methods can assist in utilizing non-invasive data. Deep learning techniques hold particular promise for reducing healthcare costs in disease diagnosis and treatment. The ability of deep learning-based methods to handle complex data and capture subtle changes is crucial for the early detection of indicators for cognitive impairment.

This review paper explores recent research articles that primarily utilized or developed deep learning methods for detecting cognitive impairment using various non-invasive data modalities (e.g., speech and language, facial video, eye gaze, and motoric mobility). Despite the availability of review papers on deep learning methods for detecting cognitive impairment~\cite{khan2021machine, khojaste2022deep}, to the best of our knowledge, none specifically discuss the use of various \textit{non-invasive} data modalities. Hence, we focused on studies that employed deep learning-based methods for non-invasive cognitive impairment detection. The main criteria for including research articles in this review are the data modalities used and the adaptation of deep learning methods in feature extraction or detection decisions.

We begin by discussing the advantages and limitations of different integrated non-invasive modalities. We then review studies that utilized various modalities to propose cognitive impairment detection systems. Ultimately, we provide a comprehensive outline of existing trends and future directions in deep learning methods for cognitive impairment detection, concluding with the overall impact of these methods on early detection of cognitive decline. This knowledge can lead to better insights into disease progression across large populations.

The organization of this review paper is as follows. Sec.~\ref{sec.Extracted Features} discusses non-invasive data as indicators of cognitive decline from a medical perspective and elaborates on the rationale behind using deep learning for analyzing such data. Sec.~\ref{sec.Datasets} introduces the datasets containing modalities captured through non-invasive techniques. Sec.~\ref{sec.Modeling and Modalities} reviews articles that utilized various modalities of non-invasive data to detect cognitive impairment. Sect.~\ref{sec.Performance} presents the evaluation performances of the reviewed studies. Sec.~\ref{sec.Challenges} discusses the challenges and suggests future research directions. Finally, Sec.~\ref{sec.Conclusion} concludes the paper with a summary of our thoughts and findings.

\begin{figure*}
    \centering
    \includegraphics[width=\linewidth]{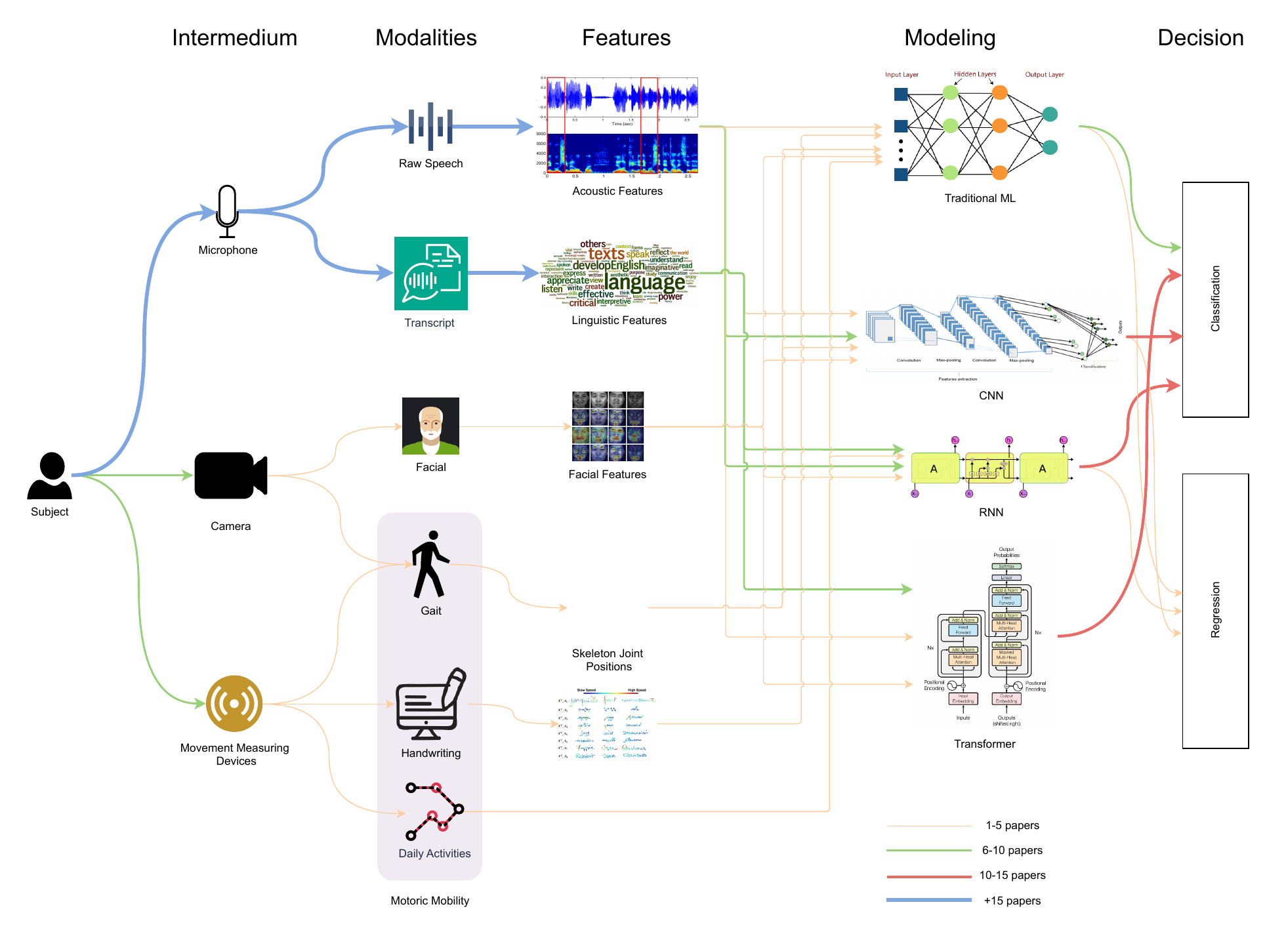}
    \caption{An overview of the reviewed papers (better viewed in colors)}
    \label{fig.overview}
\end{figure*}

\section{Cognitive Impairment Indicators} \label{sec.Extracted Features}
In this section, we investigate the medically supported motivations behind using diverse non-invasive modalities and their features for detecting cognitive status, leveraging deep learning algorithms. Specifically, we focus on \textit{Speech \& Language}, \textit{Facial}, and \textit{Motoric Mobility} indicators.

Progressive neurodegenerative cognitive conditions encompass a spectrum of cognitive decline stages, beginning with prodromal disease, and progressing through subjective cognitive impairment, Mild Cognitive Impairment (MCI) and progressing to dementia~\cite{better2024alzheimer}. The cognitive stage of patients can be assessed by examining certain predetermined factors while they engage in specific tasks (i.e., answer questions, and perform executive functions). This evaluation process involves observing how an individual performs and responds during these tasks, which helps to identify their cognitive stage. By carefully analyzing their behavior and reactions, it is possible to determine their cognitive development stage, based on those established evaluation criteria. Determining an individual's cognitive status offers valuable insight into the current challenges and guides the implementation of appropriate interventions.

At present, there exists no cure for neurodegenerative cognitive impairment. However, several therapies and lifestyle interventions, depending on the cause and the stages of the impairment, may slow the cognitive decline and improve patient and caregiver quality of life~\cite{better2024alzheimer}. For example, the Food and Drug Administration (FDA) has approved two IV medications that selectively act on the beta-amyloid protein associated with Alzheimer's disease that significantly reduces clinical decline on multiple scales, but would have no effect on cognitive impairment associated with other proteins~\cite{cummings2023lecanemab}. Typically, an evaluation of cognitive impairment begins when the patient or others who know the patient well describe changes in everyday life associated with these changes. In addition to gathering personal history, the clinician will typically perform or order cognitive testing to be done that better describes the severity and nature of the cognitive change. Depending on the findings of these initial tests, medical imaging may be ordered, such as Computed tomography (CT), magnetic resonance imaging (MRI), or positron emission tomography (PET). Blood tests and sometimes tests of cerebrospinal fluid (CSF) may also be ordered. 

Through this process, medical providers further specify which aspect of cognition (i.e., cognitive domain) is primarily impaired.  Examples may include disorders of language (aphasia), memory (amnesia), planning and attention (executive dysfunction), knowledge (agnosia), and more. These assessments implicate specific brain regions affected, such as the medial temporal lobes for memory processing and the parietal lobes for sensory information and spatial awareness~\cite{arciniegas2013behavioral}. 

While these methods can provide valuable insights into the patient's cognitive status, they require time and expertise to administer. An aging population creates a greater demand for evaluations than can be readily met, leading to delays in assessments and, consequently, delays in appropriate treatments and recommendations.

Deep learning applications have expanded to detect subtle behavioral and cognitive ability changes in cognitively impaired patients using non-invasively collected data modalities, such as speech, facial expressions, and motoric mobility. In the following, we discuss three primary non-invasive indicators of cognitive impairments, paralleling the medical explanations, focusing on speech \& language, facial, and motoric mobility modalities.

\subsubsection{Speech Indicators} 
Speech is the most extensively studied modality in deep learning for detecting cognitive status, primarily due to the early availability of public datasets such as the Pitt Corpus~\cite{becker1994natural}. These datasets have enabled researchers to develop various methods for identifying cognitive impairment at different stages. Additionally, studies have shown that both \textit{acoustic} and \textit{linguistic} features are crucial in identifying pathology characteristics. Acoustic features, signal representations related to sound perception, are theoretically language-agnostic, although variations in sounds and pronunciations across different languages and accents can affect them. On the other hand, linguistic features are derived from the meaning and understanding of the speech. These features encompass aspects such as syntax, semantics, and pragmatics, which are intrinsic to a specific language. While acoustic features capture how something is said, linguistic features focus on what is being said, providing deeper insights into the speaker's intent, context, and the conveyed message.

From a technical perspective, the analysis of human speech can be categorized into acoustic features (i.e., formats, pitch, and phonemes) and linguistic features (i.e., morphemes, words, phrases/sentences, and contextual meaning). Speech generation, which involves physical mechanisms and linguistic output, provides rich indicators for detecting cognitive impairment. However, these abnormalities are not directly observable by human perception and require technical integrations for detection. We will review acoustic and linguistic characteristics and features in the following.

\textbf{Acoustic Perspective:} Human speech includes formats, a local maxima in the speech spectrum, which define the acoustic resonance of the vocal tract. Voiced sounds emerge as air traverses the vocal tract, with the brain controlling various muscles involved in sound generation. 

The features embedded within voices possess the potential to disclose the speaker's medical condition. To illustrate, studies have linked specific acoustic features to cognitive declines in speech signals, including vocal cord deficiencies~\cite{mugler2018differential}. However, it is not a straightforward indication of the individual's medical condition, especially for older adults. To clarify, seniors usually exhibit changes in acoustic characteristics after age 60, due to a combination of cognitive and more peripheral changes~\cite{mahon2022voice, galluzzi2018aging, vaca2015aging}. Likewise, the baseline pitch often varies based on the speaker's gender. 

As a result, pre-trained deep machine learning models such as wav2vec2.0~\cite{baevski2020wav2vec} and VGGish~\cite{hershey2017cnn} for extracting acoustic features might introduce bias stemming from the general population. Therefore, it is crucial to analyze these features in conjunction with other conditions, taking into account appropriate adjustments for age-related changes. This is achievable by leveraging recent deep learning methods that are proven to be effective at processing complex and large datasets including speech. These methods enable the analysis of acoustic features to detect anomalies that are imperceptible to the human ear.

Many research studies have utilized predefined acoustic features versus those that investigated the use of raw speech data with deep learning methods to detect cognitive status. Predefined acoustic features are interpretable by researchers as related to cognitive decline. However, subtle nuances of speech may be overlooked. Whereas, utilizing raw speech with deep learning models can increase the modeling complexity and potentially reveal novel patterns in the speech.

\textbf{Linguistic Perspective:} Words are structured into sentences that express thoughts within speech. From a medical perspective, the analysis of linguistic patterns can detect deviations in language usage that suggest cognitive decline~\cite{verfaillie2018more}. Different aspects of speech relate to different aspects of cognition, supported by different anatomical substrates and compromised by different disease processes. For example, complex grammatical construction is subserved by the left frontal operculum. When this area is compromised it can result in grammatically simplified or incorrect phrasing, as is related to the nonfluent variant of primary progressive aphasia (nfvPPA), a neurodegenerative condition most commonly related to an abnormal convolution of a protein called tau~\cite{mesulam2001primary}. By identifying such grammatical errors in speech, then, an algorithm may implicate a brain region, a neurological syndrome, and predict histopathological findings under a microscope.

While conditions such as nfvPPA are rare and specific to language, subtle changes such as increased pause length and changes in linguistic coherence may signify the presence of more common conditions such as Alzheimer’s disease. Several groups worldwide have demonstrated some ability to predict MCI or Alzheimer’s disease~\cite{petersen2016mild}; however, more work remains to be done to ensure the results are reproducible across different cultures, languages, and populations.

Due to the power of Large Language Models (LLMs) in natural language processing and understanding, they are widely used to detect cognitive impairment by analyzing linguistic features such as grammar and word choice. While these features may vary with personal knowledge and experience, deep learning models can leverage extensive speech datasets to identify patterns that indicate cognitive impairment across diverse populations.

\textbf{Merging Perspectives:} Merging speech and language approach allows for a comprehensive assessment of cognitive health based on speech, providing valuable insights for early detection and monitoring. To illustrate, while both the acoustic and linguistic aspects of speech generation depend on the brain, they offer different perspectives on speech analysis, ensuring gaining a more comprehensive understanding of cognitive health.

\subsubsection{Facial Indicators}

Interactions between clinical personnel and patients are crucial for detecting cognitive impairment. As part of this delicate act of clinical communication, careful interpretation of paralinguistic as well as linguistic signals is required. Facial expressions, whether intentional or not, are emphasized in tasks involving affective reception and expression, as the neurodegenerative process may disrupt the physiological linkage between subjectively experienced and expressed emotion~\cite{pressman2023incongruences}. Those with neurodegenerative conditions, particularly forms of behavioral variant frontotemporal dementia (bvFTD) also struggle to interpret the facial expressions of others accurately~\cite{pressman2014diagnosis}, and similar difficulties are also reported in conditions as common as Alzheimer's disease, albeit to a lesser extent~\cite{pressman2017relative}. 

In studying these issues, it is important to distinguish between facial expressions originating from natural (spontaneous) reactions and those elicited by stimuli when assessing cognitively impaired subjects. As cognitive impairments can severely affect comprehension of facial expressions, utilizing computer vision techniques to extract facial features can be a fundamental step for detecting cognitive impairment using facial indicators. Recent studies have investigated the effectiveness of facial indicators through the development of several models that extract explainable facial features such as head poses, facial expressions, and Facial Action Coding System (FACS), occurring either intentionally or unintentionally~\cite{fei_novel_2022}. 

\subsubsection{Motoric Mobility Indicators} 
Many forms of neurodegenerative conditions that impact cognition also impair motor control~\cite{katz2016early}. Physical challenges with normal aging observed and evaluated in clinical settings are often linked to cognitive status~\cite{harada2013normal}. In addition to changes of healthy aging as well as motor controls more common in specific conditions such as Parkinson’s disease, many neurodegenerative conditions may ultimately involve loss of motor control in addition to cognitive changes. Furthermore, many cognitive processes such as processing speed or visuospatial awareness may become apparent in movement measures.

Studies have shown variations in pressure, stroke, and speed in handwriting, as well as alterations in drawing patterns, might reflect the early stages of cognitive impairments impacting frontal, occipital, or parietal lobes~\cite{rentz2021association}. Some drawing tasks are designed to evaluate the precision of vertical and horizontal lines within these shapes. These abnormalities are considered during neuropsychological assessments, where subjects are asked to replicate various shapes~\cite{nasreddine2005montreal}. 

Recent advancements in wearable and non-wearable measuring devices, allow for the continuous, non-intrusive monitoring of physical movements. Such data collection supports ongoing monitoring and early detection of cognitive changes. Ultimately, deep machine learning algorithms can analyze this captured data to identify abnormal movement patterns indicative of neurological, musculoskeletal, or other issues. Recently, using deep learning methods to detect cognitive conditions has become increasingly popular. These methods analyze different activities such as handwriting, drawing, and walking to identify early symptoms of cognitive impairment~\cite{cilia_handwriting-based_2021}.

\subsubsection{Multi-modal Indicators}
Studies have investigated how combining various modalities may enhance the performance of detecting and longitudinal assessment of cognitive impairment~\cite{poor2024multimodal}. Integrating different modalities has the potential to improve detection performance by capturing more indicators, and introducing different combinations of indicators that may be more powerful than an individual indicator in isolation. However, these integrations and interactions also increase the complexity of the technical models and data management. To the best of our knowledge, there is currently no established framework for fusing different modalities, particularly when considering the temporal aspects of features.

With that in mind, AI techniques can enhance the detection of cognitive impairments in clinical settings, resulting in more effective and accurate diagnoses. This improvement in precision and sensitivity can significantly enhance overall performance.

\section{Datasets} \label{sec.Datasets}

In this section, we review some existing datasets used in studies for detecting MCI and AD/ADRD using non-invasive data collection methods. We categorize the datasets based on the type of captured data from the individuals as follows: 1- \textbf{Speech-based datasets}, containing the speech recordings and/or transcripts of human subjects. 2- \textbf{Visual-based datasets}, capturing the human subjects' visual appearance such as facial. 3- \textbf{Movement-measuring-based datasets}, capturing individuals' behaviors using wearable and non-wearable measuring sensors while performing a specific task. 4- \textbf{Multi-modal datasets}, presenting more than one data modality of the individuals, including speech, movement, or visual appearance. 

We explore the datasets and provide a brief introduction for each category based on their publication year to ensure a consistent representation. While the primary focus of this section is on publicly available datasets, we also briefly discuss private datasets at the end of each category. The private datasets are collected for a specific study and usually are not available to the research community. 

\subsection{Speech-based Datasets}
In this section, we review the existing speech-based datasets in chronological order of their release. Some studies have considered speech modality out of multimodal datasets. These datasets are introduced later as multimodal datasets in Sec.~\ref{subsec.Dataset.multi_modal}.  

\subsubsection{Pitt Corpus} 
Pitt Corpus is a repository of audio recordings, including longitudinal neuropsychological assessments conducted at the Pittsburgh University School of Medicine~\cite{becker1994natural}. The dataset required participants' responses to four tasks including the Cookie Theft photo description, a Word Fluency task, Story Recall, and Sentence Construction. Participants in the study were screened for cognitive impairment and thereby, they were labeled into two conditions: Control and AD patients. Overall, Pitt Corpus comprises 101 individuals for the Healthy Control (HC) group aged 46.2-81.9 with balanced gender representation and 181 AD patients aged 50-88.7 but with two-thirds females. The dataset is publicly available with manual transcripts of the speech.

\subsubsection{WLS Dataset}
Wisconsin Longitudinal Study (WLS) is a long-term and large-scale study of random graduates from Wisconsin high schools in 1957. Each participant completed six surveys spanning from 1957 to 2011. Notably, the surveys in 2004 and 2011 included responses to cognitive tasks~\cite{herd2014cohort}. The participants' responses to the cognitive tasks were audio-recorded, with a total number of 1264 in 2004 and 1370 in 2011. To be more specific, in the 2011 survey, the participants were asked to describe the Cookie Theft image.

Despite the extensive volume of the data and long-term nature of WLS research, this dataset has two main issues: 1- A large majority of the recorded responses belong to HC participants, exacerbating class imbalance issues for deep machine learning applications. 2- The dataset lacks clinical cognitive impairment scores although these may still be inferred from linguistic cognitive tests.

\subsubsection{ADReSS}
Alzheimer’s Dementia Recognition through Spontaneous Speech (ADReSS) Challenge is a publicly introduced dataset for cognitive impairment detection challenge at INTERSPEECH 2020~\cite{luz_alzheimers_2020}. The dataset contains audio recordings and manual transcripts of participants describing the Cookie Theft picture. To be more detailed, the ADReSS dataset comprises two categories: healthy controls and Alzheimer's, with each group consisting of 156 participants. The participants' ages range from 50 to 80 years, with a balanced representation of both genders in both groups.

The ADReSS dataset includes the AD labels along with the Mini-Mental State Examination (MMSE) scores of the participants. These labels and scores help in developing classification and regression systems. Furthermore, the dataset presents training and test subsets to ensure a standardized evaluation of methods.

\subsubsection{ADReSSo}
Alzheimer’s Dementia Recognition through Spontaneous Speech \textit{only} (ADReSSo) is a speech-based Challenge introduced at INTERSPEECH 2021~\cite{luz_detecting_2021}. Specifically, the ADReSSo provides two distinct datasets: the prognostic dataset and the diagnostic dataset. The prognostic dataset contains audio recordings of AD patients performing a semantic fluency task. Whereas, the diagnostic dataset includes audio recordings of AD patients and NC individuals describing the Cookie Theft picture. Notably, compared to ADReSS, the ADReSSo included the prognostic dataset to serve as a baseline for cohort study. Additionally, the dataset contains only audio recordings of the participants' responses.

Similar to the ADReSS dataset, the ADReSSo's diagnostic dataset provides AD labels and MMSE scores. On the contrary, the ADReSSo's prognostic dataset serves as a predictor of the progression of cognitive decline in cognitively impaired individuals over time.

\subsubsection{NCMMSC2021}
National Conference on Man-Machine Speech Communication 2021 Alzheimer’s Disease Recognition Challenge (NCMMSC2021) is a public dataset for the AD Recognition Challenge at the NCMMSC 2021. The dataset is in Mandarin which includes speech recordings and corresponding transcripts from subjects participating in three tasks: picture description, fluency test, and free conversation during interviews. Overall, this dataset contains 280 participants as follows: 26 AD patients, 53 MCIs, and 44 NCs~\cite{ying_multimodal_2023}.

\subsubsection{Private datasets}

Some studies have collected their data domestically for various reasons, although this process can be costly and unreasonable for a specific study. In the following, we briefly describe the data-collecting process of some of the speech-based studies.

Nishikawa~\etal~\cite{nishikawa_detecting_2021} recruited 30 older adults to take speech recordings of participants and then transcribed the speech. The participants took the MMSE to evaluate their cognitive status. They split the subjects into the MCI and NC groups by referring to the MMSE scores. Subjects with the MMSE score range of [23, 27] are labeled as MCI, while the resting ones fall into the NC group. There are 15 MCIs and 15 NCs. After removing the silence section of the speech data, they cut the data into 3-second clips with a sampling frequency of 48~kHz for further study.

\subsection{Visual Datasets}

Visual datasets are collections of images or videos that capture non-invasive indicators of cognitive impairment, such as variations in participants' facial or body appearance. Unlike other speech-based datasets, almost all visual datasets are self-collected and created for specific studies. It is important to note that some datasets used for visual data also contain other modalities. Therefore, we will introduce them as multi-modal datasets in Sec.~\ref{subsec.Dataset.multi_modal}.

\subsubsection{Private datasets}

\textbf{Gait data} encompass the body motions. These motions can be captured using various ways; however, here, we focus on gait captured by visual devices.

You~\etal~\cite{you_alzheimers_2020} collected skeleton joint positions from 35 NC, 35 individuals with MCI, and 17 with AD, in a lab environment, using Microsoft Kinect V2.0 cameras. In detail, they deployed 8 and 6 devices in the Neurology and Geriatrics Departments, respectively, and set the tilt angle as 27 degrees. Additionally, electroencephalogram (EEG) data were acquired with patients' eyes closed and open for 8 minutes each.  

You~\etal~\cite{you_alzheimers_2021} conducted a study by collecting gait data from 53 individuals with MCI or AD and 35 in the control group, both aged over 46 years old. The data collection took place at Keio University School of Medicine, utilizing Kinect cameras. Participants were instructed to walk the length and back between two devices positioned 10 meters apart. 

\textbf{Eye gaze data} is a projection of where a person is looking collected using visual recording devices. For instance, Zuo~\etal~\cite{zuo_deep_2023} collected eye-tracking data for cognitive impairment classification, including a total of 106 participants, consisting of 38 with AD and 68 classified as normal, at the Tianjin Huanhu Hospital, Tianjin, China. They employed a non-invasive eye-tracking system with stereo stimuli to record and analyze visual reactions associated with various emotions through gaze tracking.

\begin{table*} [ht]
\centering
\caption{The general information of the discussed datasets.}
\resizebox{\textwidth}{!}{
\begin{tabular}{llllll}
\hline
 Dataset & Longitudinal & Data Modality & Language & Conditions Distribution & Studies \\
\hline   

Pitt~\cite{becker1994natural} & Yes & Speech, Text & English & 101 HC and 181 AD  & \cite{khan_stacked_2022,orimaye_deep_2018,fritsch_automatic_2019} \\
\multirow{2}{3cm}{WLS~\cite{herd2014cohort}} & \multirow{2}{1cm}{Yes} & \multirow{2}{2.cm}{Speech} & \multirow{2}{1cm}{English} & \multirow{2}{2cm}{} & \multirow{2}{*}{\cite{guo_crossing_2021}}\\
 &  &  &  &  \\ 
ADReSS~\cite{luz_alzheimers_2020} & No & Speech, Text & English & 156 HC and 156 AD & \cite{cummins_comparison_2020,guo_crossing_2021,koo_exploiting_2020,syed_automated_2021} \\
\multirow{2}{3cm}{ADReSSo~\cite{luz_detecting_2021}} & \multirow{2}{1.cm}{No} & \multirow{2}{2.cm}{Speech} & \multirow{2}{1.cm}{English} & \multirow{2}{2.cm}{} & \cite{pan_using_2021, gauder_alzheimer_2021,wang_modular_2021} \\
 &  &  &  & & \\ 
NCMMSC2021  & No & Speech, Text & Chinese & 26 AD, 53 MCI, and 44 NC & \cite{ying_multimodal_2023} \\
 
&  &  &  &  & \\ 
\multirow{2}{3cm}{PROMPT~\cite{kishimoto_project_2020}} & \multirow{2}{1cm}{Yes} & \multirow{2}{2.cm}{Speech, Text, Video, Biosignal} & \multirow{2}{1cm}{Japanese} & \multirow{2}{2.cm}{} & \cite{rodrigues_makiuchi_speech_2021, zheng_detecting_2023} \\
&  &  &  & & \\ 
\multirow{2}{3cm}{I-CONECT~\cite{dodge2024internet}} & \multirow{2}{1cm}{Yes} & \multirow{2}{2.cm}{Speech, Text, Video} & \multirow{2}{1cm}{English} & \multirow{2}{2cm}{34 MCI and 34 NC} & \cite{sun_mc-vivit_2024, fard2024linguistic, alsuhaibani2024mild} \\
&  &  &  & & \\

\hline
\end{tabular}}
\label{table.datasets_summary}
\end{table*}

\subsection{Movement-measuring-based Datasets} \label{subsec.dataset.sensor-based}

In this section, we review the existing Movement-measuring-based datasets in chronological order. Unlike speech-based datasets, most movement-measuring-based datasets are private to certain research groups.

\subsubsection{Private datasets}
While these datasets might not be as comprehensive as the publicly available datasets, reviewing such datasets gives readers a wider overview of how using measuring movement data might contribute to the detection of AD.  

\textbf{Gait Data:} Bringas~\etal~\cite{bringas_alzheimers_2020} conducted a private study involving 35 patients diagnosed with AD from the AFAC daycare center of Santander, Spain. Patient mobility data were gathered using accelerometer sensors from smartphones. The dataset comprised 6-hour durations for each of the 35 patients, representing various stages of Alzheimer's disease: 7 in the early stage, 18 in the moderate stage, and 10 in the severe stage. 

Aoki~\etal~\cite{aoki_early_2019} utilized a private dataset and MMSE alongside a Kinect sensor to capture whole-body movements for cognitive assessment. The study introduced a dual-task paradigm, combining movement analysis with a cognitive task (counting down from 100 by ones) to explore distinguishing gait features between healthy and cognitively impaired older adults. 

Shahzad~\etal~\cite{shahzad_automated_2022} conducted a study on gait biomarkers for MCI screening, analyzing human walking patterns during cognitive tasks. Using data from inertial sensors with triaxial accelerometers and gyroscopes on a 10-meter walkway, the study incorporated cognitive tasks such as down-counting and naming animals. 

Ghoraani~\etal~\cite{ghoraani_detection_2021} conducted a study involving 78 participants, including 32 healthy individuals, 26 with MCI, and 20 with AD, where a Zenomat system from ProtoKinetics LLC and a GAITRite system from CIR Systems, PA where utilized. Additionally, gait features were extracted using the ProtoKinetics Movement Analysis Software (PKMAS) for the Zenomat system and the GAITRite software for the GAITRite system. 

\textbf{Daily-life activities:} Narasimhan~\etal~\cite{narasimhan_early_2020} employed non-wearable sensors installed in the residences of older adults to capture data related to sleep duration, cooking time, and walking speed simulating the longitudinal activity trend for 10 older adults, incorporating 4 assessment time points per individual.

\textbf{Writing-Drawing}-based activities, including copying text, loop series, and drawings, are another type of activity often used in research on the detection of cognitive impairment. El-Yacoubi~\etal~\cite{el-yacoubi_aging_2019} gathered handwriting data from three groups, totaling 144 participants aged over 60, at a hospital in Paris. The sampling rate employed was 125Hz, with the dataset comprising x and y positions, pressure, and in-air trajectory up to 2 cm. The participants undertook seven tasks, including copying text, loop series, and drawings. Cilia~\etal~\cite{cilia_handwriting-based_2021} conducted six handwriting tasks at an Italian hospital, involving a total of 181 participants. Among them, 90 were identified as cognitively impaired, while the remaining 91 served as healthy controls.

\subsection{Multi-Modal Datasets} \label{subsec.Dataset.multi_modal}
Multi-modal datasets present more than one type of participant data, usually including speech, sensor data, or visual appearance. In the following, we review some of the public multi-modal datasets available for non-invasive recognition of AD.

\subsubsection{PROMPT Dataset}
Project for Objective Measures using Computational Psychiatry Technology (PROMPT) Dataset is a public Japanese dataset collected by Keio Medical School~\cite{kishimoto_project_2020}, divided into Dementia patients, Bipolar disorder patients, Depression patients, and NC. Each subject underwent three language tasks including free talk, questions and answers, and picture description tasks while recording subjects' speech and video during the interview. The PROMPT dataset contains facial expressions, body movements, speech, and daily life activity from subjects, and supports various research plans. 

\subsubsection{I-CONECT Dataset}
Internet-Based Conversational Engagement Clinical Trial (I-CONECT; NCT02871921) dataset is a longitudinal study of 187 socially isolated older adults. The clinical trial randomly split the participants into two categories: control and experimental groups. The experimental group underwent 30-minute semi-structured interviews four times a week for six months and then twice a week for six more months, whereas the control group received weekly checkup phone calls~\cite{yu_internet-based_2021}. Moreover, The participants are assessed three times during the study (baseline, 6 month, and 12 month) with the MoCA score and a label of MCI or NC. 

The dataset contains audio-video recordings of the interviews that were conducted using user-friendly devices (i.e., webcams). In addition, the phone check-ups are recorded. The videos capture the faces of the participants and record their responses to the interviewers. Particularly, the participants are over the age of 75 and live in either Portland (Oregon), Atlanta (Georgia), or Detroit (Michigan) in the United States. The experimental group consists of a total of 68 individuals, with an equal balance of cognitive conditions, specifically 34 each from MCI and NC. Although the dataset representations show diversified gender, they fall short in terms of racial and ethnic variety~\cite{dodge2024internet}.

Table~\ref{table.datasets_summary} summarized the general information of the datasets discussed in Sec~\ref{sec.Datasets}. We should emphasize that those datasets are not the only publicly available datasets of non-invasive data. However, researchers have developed DL models utilizing those datasets.

\section{Modeling and Data Modalities} \label{sec.Modeling and Modalities}

In this section, we review common machine-learning models utilized for detecting cognitive impairment and related conditions. We categorize these models into \textit{traditional machine learning models} and \textit{deep machine learning models}. While numerous studies have applied traditional machine learning models to cognitive impairment recognition tasks, we focused on studies that either directly used deep learning models or compared their performance with traditional machine learning approaches for building detection systems.

In \textbf{traditional machine learning} methods, feature extraction is conducted separately from classification or recognition tasks. This means that, based on the data modality, specific features are extracted first, independently of the learning task. Once features are extracted, classifiers such as K-nearest neighbors (kNN), Support Vector Machines (SVM), Random Forests (RF), or Multilayer Perceptrons (MLP), among others, are trained. These traditional machine learning models are widely adopted in the field due to their straightforward implementation via open-source software libraries.

In \textbf{deep machine learning} approaches, an end-to-end system is often built in which a neural network model learns and extracts discriminative features suitable for a particular task. These features are learned directly from the data during the training process.

One of the commonly used deep learning architectures is Convolutional Neural Networks (CNNs), which are particularly effective at extracting and capturing hierarchical features from images. CNNs extract low-level features through initial layers and progressively capture high-level features through deeper layers. CNN architectures such as ResNet~\cite{he2016deep}, VGG~\cite{simonyan2014very}, and MobileNetV2~\cite{sandler2018mobilenetv2} are commonly used models for various tasks.

Recurrent Neural Networks (RNNs) are another type of neural network that consider the sequential nature of data. They integrate input data across time steps, making them suitable for time series or sequential data. The two main architectures of RNNs are Gated Recurrent Units (GRU)~\cite{cho2014learning} and Long Short-Term Memory (LSTM)~\cite{hochreiter1997long} networks. These architectures can process sequential data in both forward and backward directions, providing a more comprehensive understanding of the sequences.

Transformers, developed in 2017 \cite{vaswani2017attention}, revolutionized many fields, especially NLP, by adopting a sequence-to-sequence approach with an attention mechanism. Initially introduced for machine translation~\cite{vaswani2017attention}, Transformers have since become useful in various applications including vision tasks. Notable Transformer models include Generative Pre-trained Transformer (GPT)~\cite{radford2019language}, Bidirectional Encoder Representations from Transformers (BERT)~\cite{devlin2018bert} for NLP, and Vision Transformers (ViT)~\cite{dosovitskiy2020image} for computer vision. The Transformer encoder is mainly used for feature extraction and classification whereas the Transformer decoder is utilized for data generation.

\subsection{Speech-based Modality} \label{subsec.Speech-based} 

Speech has emerged as a prominent non-invasive modality for detecting cognitive impairment. The appeal of speech-based methods lies in the data availability and cost-effectiveness.

\subsubsection{Acoustic-based} \label{subsec.Acoustic-based}
%
\begin{table*}[ht]
\centering
\caption{A summary of the reviewed studies that utilized acoustic features}
\label{table.acoustic_summary}
\rowcolors{2}{gray!10}{white}
\resizebox{\textwidth}{!}{


\begin{tabular}{cccccccc}
\hline
\rowcolor{gray!30}
 &  &  &                  &   & \multicolumn{3}{c}{Evaluation$^*$ (\%)}  \\ \cline{6-8} 
\rowcolor{gray!30}
\multirow{-2}{*}{Authors (Year)} & \multirow{-2}{*}{Dataset} & \multirow{-2}{*}{Language} & \multirow{-2}{*}{Features}                 & \multirow{-2}{*}{Classification Models}      & \multicolumn{1}{l}{ACC} & F1   & AUC \\ \hline
Kumar~\etal (2022)~\cite{kumar_dementia_2022}                    &    &   &                      & RF                                                                        & 87.6                    & 87.5 &  -   \\
Liu~\etal (2021)~\cite{liu_detecting_2021}             &      \multirow{-2}{*}{Pitt}          &                           &                       & CNN-Bi-LSTM-attention                                                    & 82.6                    & 82.9 &   -  \\ 
Meghanani~\etal (2021)~\cite{meghanani_exploration_2021}                & ADReSS                   &      \multirow{-3}{*}{English}                   &      \multirow{-3}{*}{Spectral Acoustic Features}      & CNN-LSTM                                                              & 64.7                    & - &  -   \\ 
Rodrigues Makiuchi~\etal (2021)~\cite{rodrigues_makiuchi_speech_2021}       & PROMPT                   & Japanese                  &   &                                                     & 80.8                    &   -   &   -  \\
Warnita~\etal (2018)~\cite{warnita_detecting_2018}                  & Pitt                     &   &                                           &                                           \multirow{-2}{*}{GCNN}                                & 73.6                    &   -   &   -  \\
Syed~\etal (2020)~\cite{syed_static_2020}                     & ADReSS                   &        \multirow{-2}{*}{English}                   &                                           & Bi-LSTM                                                                   & 74.6                    &   -   &    - \\
Nishikawa~\etal (2021)~\cite{nishikawa_detecting_2021}                & private                  & Japanese                  &     \multirow{-4}{*}{Acoustic Feature Sets}      & CNN-LSTM                                                                  & 90.8                    & 89.7 & 91  \\ 
Chlasta and Wolk (2021)~\cite{chlasta_towards_2021}         & ADReSS                   &   &           & DemCNN                                                                     & 63.6                   & 69.2 &   -  \\
Gauder~\etal (2021)~\cite{gauder_alzheimer_2021}     & ADReSSo                  &     \multirow{-2}{*}{English}  &       \multirow{-2}{*}{DL Audio Models}      &   CNN      & 78.9                    &   -   &  -   \\ 
Nishikawa~\etal (2022)~\cite{nishikawa_machine_2022}                & private                  & Japanese                  &       &         & 89.4                   & 84.4 &  -   \\

Pranav~\etal (2023)~\cite{pranav_early_2023}                   &    &    &                                           &             \multirow{-2}{*}{ViT}      & 85.7                    &  92.3    &  -   \\
Bertini~\etal (2022)~\cite{bertini_automatic_2022}              &     \multirow{-2}{*}{Pitt}       &  \multirow{-2}{*}{English}                         &                                           &                                       & 90.7                    & 88.5 &  -   \\
Berini~\etal (2021)~\cite{bertini_automatic_2021}                   &                   & Italian                   & \multirow{-4}{*}{log Mel-spectrogram}                                          &                               \multirow{-2}{*}{Autoencoder + MLP}            & 90.6                    & 90.7 &  -   \\
Pan~\etal (2020)~\cite{pan_acoustic_2020}                      & \multirow{-2}{*}{private}                      & English                   & Raw Speech                                & SincNet, Bi-LSTM, Attention layer                                          &    -                     & 87.3 &   -  \\ \hline
\rowcolor{white}
\multicolumn{4}{l}{$*$ Accuracy (ACC), Area Under the Curve (AUC)}                                           &                                        &                                    & \multicolumn{1}{l}{}    &          
\end{tabular}}
\end{table*}

Speech is usually recorded using mono or stereo microphones with various sampling frequencies (e.g., 16~kHz, 22~kHz, and 44.1~kHz) and stored as raw audio (wave files) or compressed (mp3 files). Then acoustic features are extracted from the signal waveforms. Researchers have taken various approaches to extract features that best represent the speech data. In the following, we discuss data preprocessing and feature extraction techniques, then review studies based on the adopted machine-learning techniques.

\paragraph{Data Preprocessing and Feature Extraction} \label{subsubsec.acoustic.preprocessingAndFeatures}

Effective data preprocessing and feature extraction steps are crucial for enhancing the performance of cognitive impairment detection models using speech signals. Feature extraction for acoustic data involves deriving informative components from the preprocessed speech signals.

\textbf{Preprocessing} of speech signals is essential across various studies. Several studies have preprocessed the speech signals by implementing noise removal and amplitude enhancement mehods~\cite{kumar_dementia_2022}. Most studies processed speech signals by splitting audio recordings into shorter frames~\cite{xue_detection_2021, nishikawa_detecting_2021}.

\textbf{Spectral acoustic features} are descriptions of energy distribution across speech frequencies during a specific time. Notably, these features carry physical information about the speech. For example, Mel-Frequency Cepstral Coefficients (MFCC) are extensively used because of their efficacy in capturing spectral dynamics~\cite{nishikawa_detecting_2021, nishikawa_machine_2022, kumar_dementia_2022, zolnoori_adscreen_2023}. MFCCs often combine with other features such as jitter, shimmer, fundamental frequency, formants, noise-to-harmonic ratio (HNR), and gammatone cepstral coefficients (GTCC). Log-Mel spectrograms are another spectral feature that has a graphical representation of time and spectrum of the speech~\cite{pranav_early_2023, nishikawa_machine_2022, meghanani_exploration_2021, bertini_automatic_2021, bertini_automatic_2022, lin_automatic_2022}.

\textbf{Predefined feature sets} are introduced to capture various useful information of the speech signals. These feature sets are mainly presented as challenges at the interspeech conference. For example, the Interspeech 2009 (IS09) emotion challenge feature set is proposed to show the emotional information of the speech. Sequential challenges at the Interspeech presented extended and other feature sets for paralinguistics analysis of the speech such as IS10 and ComParE. Studies have extracted these features for cognitive impairment detection~\cite{warnita_detecting_2018, syed_static_2020, rodrigues_makiuchi_speech_2021, syed_automated_2020, syed_automated_2021, ying_multimodal_2023, wang_modular_2021}.

Similarly, the Geneva Minimalistic Acoustic Parameter Set (GeMAPS) feature sets consist of voice parameters with an extended set and are referred to as eGeMAPS~\cite{eyben2015geneva}. This feature set is utilized in cognitive status detection in~\cite{syed_static_2020, gauder_alzheimer_2021, pappagari_automatic_2021}. Advanced acoustic representations like i-vectors have been applied to capture speaker-specific variations and environmental context~\cite{campbell_alzheimers_2020}. 

The availability of public libraries made these feature extraction applicable across different approaches. For instance, OpenSMILE~\cite{eyben2010opensmile} and Librosa~\cite{mcfee2015librosa} are among the leading libraries in extracting different acoustic features.

\textbf{Deep learning audio models} are advanced models that extract useful features of given speeches. These models are pre-trained on large audio datasets to represent generalized features of the speech. There are few leading models utilized for this purpose~\cite{mehrish2023review}. For example, Bag-of-Audio-Words (BoAW)~\cite{pancoast2012bag}, VGGish, wav2vec2.0~\cite{baevski2020wav2vec}, and x-vector~\cite{snyder2018x} are used in extracting different dimensions of the speech signal~\cite{gauder_alzheimer_2021, koo_exploiting_2020, cummins_comparison_2020, syed_automated_2020, wang_modular_2021,syed_automated_2021, pappagari_automatic_2021, chlasta_towards_2021, ying_multimodal_2023, pan_using_2021, li_leveraging_2023}.

\textbf{Raw speech signals} refer to the speech's waveform. This method omits the feature extraction step of the speech. Few studies adopted this approach in detecting cognitive impairment~\cite{pan_acoustic_2020}.

\textbf{Feature selection} is a method to utilize the most important features out of the selected features. The feature selection is used to reduce the vector representation of the feature vectors. There are many methods in feature selection; however, we are referring to ones used in studies that detect the cognitive status of individual speeches~\cite{nishikawa_detecting_2021}. The Mann-Whitney U-test and SVM feature selection are among the used methods in these studies.

These feature representations capture speech nuances among studied individuals, highlighting patterns that may indicate cognitive decline.

\paragraph{Modeling Techniques}

The choice of modeling techniques plays a significant role in the performance of cognitive impairment detection systems. These techniques can be broadly categorized into traditional machine learning methods and deep learning methods.

\textbf{Traditional Machine Learning Methods}

Kumar~\etal~\cite{kumar_dementia_2022} compared the performance of traditional ML models such as SVM and RF against various deep learning models. They assessed the effectiveness of these models using the extracted features. Syed~\etal~\cite{syed_static_2020} analyzed dementia detection using static modeling techniques like Support Vector Classifier (SVC) and RF. They further compared these methods with dynamic modeling approaches, underscoring the superior performance of dynamic modeling for capturing acoustic speech feature variations over time.

\textbf{Deep Learning Methods}

The transition towards deep learning has been driven by its superior performance in handling complex features and large datasets. 

Warnita~\etal~\cite{warnita_detecting_2018} explored CNN and Time-Delay Neural Network (TDNN) models with a gating mechanism for AD detection. Their approach demonstrated the ability of GCNNs to capture temporal information effectively. Nishikawa~\etal~\cite{nishikawa_detecting_2021} implemented a 1D CNN-LSTM model, combining convolutional layers' feature extraction capabilities with LSTM’s temporal sequence modeling strength.

Meghanani~\etal~\cite{meghanani_exploration_2021} compared CNN-LSTM, ResNet-LSTM, and pBLSTM-CNN architectures for AD classification, highlighting the robustness of deep learning methods over traditional models. Liu~\etal~\cite{liu_detecting_2021} combined CNN layers for local context modeling with Bi-LSTM layers for global context modeling, followed by attention pooling for AD detection, demonstrating a comprehensive approach to modeling speech data. Kumar~\etal~\cite{kumar_dementia_2022} also implemented a Parallel Recurrent Convolutional Neural Network (PRCNN) to distinguish speech segments based on the speaker's cognitive condition. 

Berini~\etal~\cite{bertini_automatic_2021} implemented an Autoencoder with GRU layers as an encoder and decoder of the network. Subsequently, they implemented an MLP as a classifier. Similarly, they proposed the same architecture in a later study that used a different dataset~\cite{bertini_automatic_2022}.

Rodrigues Makiuchi~\etal~\cite{rodrigues_makiuchi_speech_2021} used Gated Convolutional Neural Networks (GCNN) for dementia detection, showing that GCNN can effectively model the correlation between low-level descriptors across time frames. Chlasta and Wolk~\cite{chlasta_towards_2021} proposed the DemCNN sequential architecture with multiple Conv1D layers, which outperformed traditional classifiers like SVM and MLP. Pranav~\etal~\cite{pranav_early_2023} demonstrated the superiority of ViT over traditional classifiers such as Random Forest when classifying Alzheimer’s detection using log-Mel spectrograms.

In summary, robust data preprocessing and feature extraction coupled with advanced modeling techniques are essential for effective cognitive impairment detection using speech data. The evolving deep learning methods, especially those integrating convolutional, recurrent, and attention architectures, have shown substantial improvements over traditional machine learning models, offering promising directions for future research and application in this domain. Table~\ref{table.acoustic_summary} summarizes the reviewed studies and includes the evaluation performance of each approach which will be discussed later.

\subsubsection{Language-based} \label{subsec.Language-based}

\begin{table*}[]
\centering
\caption{A summary of the reviewed studies that utilized linguistic features}
\label{table.linguistic_summary}
\rowcolors{2}{gray!10}{white}
\resizebox{\textwidth}{!}{

\begin{tabular}{ccccccccc}
\hline
\rowcolor{gray!30}
 &  &  &                 &     &  \multicolumn{3}{c}{Evaluation $^*$ (\%)} \\ \cline{6-8}
\rowcolor{gray!30}
\multirow{-2}{*}{Authors (Year)} & \multirow{-2}{*}{Dataset} & \multirow{-2}{*}{Language} & \multirow{-2}{*}{Features}                & \multirow{-2}{*}{Classification Model}       & ACC          & F1          & AUC         \\ \hline
Searle~\etal (2020)~\cite{searle_comparing_2020}                   &   &   &   & SVM                                                              & 81           &    -         &     -        \\
Meghanani~\etal (2021)~\cite{meghanani_recognition_2021}                &                          &                           &                                          & CNN                                                               & 83.3         &      -       &     -        \\
Yuan~\etal (2021)~\cite{yuan_pauses_2021}                     &                          &                           &                                          &                                 & 89.6         &     -        &      -       \\
Liu~\etal (2022)~\cite{lin_automatic_2022}                      &    \multirow{-4}{*}{ADReSS}    &                                                                    &                                          &                         & 88           &     -        &    -         \\
Guo~\etal (2021)~\cite{guo_crossing_2021}                      & ADReSS, WLS              &                           &                                          &                                        \multirow{-3}{*}{fine-tuning LLM}   & 97.9         &      -       & 99.2        \\
Roshanzamir~\etal (2021)~\cite{rodrigues_makiuchi_speech_2021}              &     &                           &                                          & augmentation layer with Bi-LSTM or LR                            & 88.08        &      -       &     -        \\
Fritsch~\etal (2019)~\cite{fritsch_automatic_2019}                  &  \multirow{-2}{*}{Pitt}    &     \multirow{-7}{*}{English}  &             & LSTM                                                              & 85.6         &       -      &      -       \\

Casanova~\etal (2020)~\cite{casanova_evaluating_2020}                 &  private     & Portuguese                &    & RNN model                                                        & -            & 75          &      -       \\
Liu~\etal (2022)~\cite{liu_improving_2022}                     & Pitt, ADReSS       & English/Chinese           &                                          &                              & 93.5         & 90.2        &     -        \\
Tsai~\etal (2021)~\cite{tsai_efficient_2021}                     & Pitt, NTUHV              & English/Chinese           &            &                                \multirow{-2}{*}{Transformer encoder}       & 84           &      -       & 92          \\
Chen~\etal (2019)~\cite{chen_attention-based_2019}                    &     &   &  & BiGRU, CNN, attention layer                                       & 97.4         &      -       &       -      \\
Khan~\etal (2022)~\cite{khan_stacked_2022}                     &     \multirow{-2}{*}{Pitt}           &                           &    \multirow{-12}{*}{Word Embeddings}     & parallelized (CNN, CNN+Bi-LSTM, Bi-LSTM)                          & 93.3         & 92.2        & 85.7        \\
AI-Atroshi~\etal (2022)~\cite{ai-atroshi_automated_2022}               & Private  & Hungarian                 & GMM, DBN                                 & MLP                                                              & 90.3         & 90.2        &      -       \\
Wen~\etal (2023)~\cite{wen_revealing_2023}                      &   Pitt     &                           & PoS                                      & self-attention + attention layer + CNN                            & 92.2         & 95.5        & 97.1        \\
Alkenani~\etal (2021)~\cite{alkenani_predicting_2021}                & Pitt, ADBC               &                           & lexicosyntactics, n-gram                 & several ML                                                        & -            &     -        & 98.1        \\
Wang~\etal (2021)~\cite{wang_explainable_2021}                    & Pitt                     &                           & PoS, sentence embeddings                 & C-attention                                                       & 91.5         & 94.5        & 97.7        \\
Fard~\etal (2024)~\cite{fard2024linguistic}                    & I-CONECT                 &    \multirow{-6}{*}{English}  & Sentence embeddings                      & Transformer encoder                                              & 85.2         &      -       & 84.8        \\ \hline
\rowcolor{white}
\multicolumn{4}{l}{$*$ Accuracy (ACC), Area Under the Curve (AUC)}                                                                &                                          &                                      &             &            

\end{tabular}}
\end{table*}

In recent years, we have observed a great advancement in the field of NLP. The success is mostly due to new modeling approaches such as Transformers and other algorithms for creating language models using deep neural networks and robust word embedding methods. These include pretrained LLMs such as GPT and BERT that are successfully applied in various tasks in NLP including sentence and text classification, sentiment and emotion classification to text generation, and language translation. In this section, we review the papers that utilized language as the main data modality for the detection of cognitive impairment and AD/ADRD. We first review the data preprocessing of the transcripts and then focus on modeling methods that build the detection system.

\paragraph{Data Preprocessing and feature extractions} \label{subsubsec.lang.preprocessingAndFeatures}
Researchers have applied methods to prepare and/or represent the language data. The first step in data preprocessing is transcribing the participants' speech especially when the dataset does not already include manual transcriptions. Various methods employ pretrained deep learning methods to transcribe spoken words. This preprocessing step can significantly impact the performance of the detection system. Automated Speech Recognition (ASR) is a machine-learning application that helps convert human-recorded speech into text. 

Converting the words (aka tokens) into embedded vectors is essential in language preprocessing before implementing the deep learning models. In the following, we review the well-known and commonly used word embedding algorithms.
Considering the tokenization of generating word embedding from transcripts, it has a special integration that removes stop words. Also, the performing of lemmatization ensures all the words in the transcript are valid~\cite{khan_stacked_2022}. 


\textbf{Word Embeddings} are multi-dimensional vectors that hold words’ representations by capturing useful information from meaning and contextual usage. The word embedding generation can be categorized based on their integration methods. Word embedding systems are either \textit{Count-based} or \textit{prediction-based} method. Most studies have extracted word embeddings as the initial part of their work. 

\textit{Count-based} methods represent words by their co-occurrence in a large corpus or mutual information of two words. Word2Vec~\cite{mikolov2013efficient}, GloVe~\cite{pennington2014glove}, and FastText~\cite{bojanowski2017enriching} are among the well-introduced models in the field of NLP. Meghanani~\etal~\cite{meghanani_recognition_2021} have extracted word embeddings by utilizing the GloVe model. Similarly, Chen~\etal~\cite{chen_attention-based_2019} investigated applying word embeddings from Word2Vec and GloVe. Interestingly, Khan~\etal~\cite{khan_stacked_2022} investigated the possibility of initializing the word embeddings randomly and then performing the detection of participants' cognitive conditions on the transcripts while comparing. However, word embeddings from GloVe provided the model with more reliable vectors to have an overall better detection. 

\textit{Prediction-based} methods generate the word embeddings during predicting the language tasks and update the word embedding vectors in the training process. These methods are utilized in deep learning models and gained more attention after the introduction of Transformers.

Several studies have compared the detecting algorithms by extracting the word embedding from different pretrained LLMs such as BERT~\cite{devlin2018bert}, GPT~\cite{radford2019language} and then applying the same classifier~\cite{searle_comparing_2020,roshanzamir_transformer-based_2021}. Although studies consider different datasets in their studies, they extracted the word embeddings of transcripts by utilizing the same language models~\cite{guo_crossing_2021}. Others have integrated different detecting methods while considering the same word embeddings~\cite{yuan_pauses_2021}. 

The method of implementing word embeddings goes for non-English languages as well with the consideration that the pretrained models have been trained using the targeted language. Studies have extracted word embeddings to detect the cognitive conditions of participants from different languages. For example,~\cite{casanova_evaluating_2020} have extracted word embeddings on Portuguese transcripts, whereas~\cite{tsai_efficient_2021} have utilized the fasttext Multilanguage model to extract word embeddings of Chinese and English transcripts.  


The low-level linguistic features of transcripts are not exclusive on word embeddings. Studies have integrated other features that are linked to the word representation in the transcripts. The extraction of sentence embeddings has been studied in the detection of cognitive conditions from participants' transcripts. The extraction of sentence-level representation is adopted in~\cite{wang_explainable_2021,fard2024linguistic} and part-of-speech (PoS) features~\cite{wang_explainable_2021,wen_revealing_2023}. 

The take on the hesitation and gap filler words is the consideration of speech pauses from linguistic feature perspectives. Yuan~\etal~\cite{yuan_pauses_2021} have encoded the pauses of participants in the transcripts, where the pauses exceeding 50 ms are given unique characters. Consequently, pauses under 0.5s, between 0.5s and 2s, and over 2s are assigned different characters. BERT~\cite{devlin2018bert} and ERNIE~\cite{sun2020ernie} are fine-tuned based on the updated transcripts of the participants after encoding the pauses.


Linguistic patterns are utilized to find patterns among cognitive impairment participants by calculating the lexicosyntactic feature and character n-gram spaces. Alkenani~\etal~\cite{alkenani_predicting_2021} selected the feature spaces of lexicosyntactic features by finding the correlation based on the inter-correlations and target correlations according to a certain threshold.

The perplexity values are calculated from transliterations of the participants based on both cognitive conditions as in \cite{fritsch_automatic_2019}. Similarly, Colla~\etal~\cite{colla2022semantic} have calculated the perplexity of participants' transcripts to detect cognitive impairment. 

Searle~\etal~\cite{searle_comparing_2020} have calculated the Term Frequency-Inverse Document Frequency (TF-IDF) to down-wight the common cross-document words and increase the weights of rare cross-document but frequent intra-document words. This method integrates bag-of-words (BoW) to detect cognitive impairment. However, it omits the consideration of words' sequential representation.

Saltz~\etal~\cite{saltz_dementia_2021} introduced a vocabulary size of type-token ratio (TTR) of semi-structured interview transcripts. They compared the TTR to evaluate semi-structured interview transcripts based on the cognitive conditions of participants. Several language models are integrated with BERT~\cite{devlin2018bert}, XLNet~\cite{yang2019xlnet}, and ELECTRA~\cite{clark2020electra} to extract linguistic features. 

The global contextual representation of the extracted features is desired to present an overall representation of the participants’ generated sentences. This technique is crucial especially when the word vectors are extracted either from count-based word embeddings or LLMs models. The extraction of local features is essential as they show the dependency of words’ vectors that are close to each other. Moreover, modeling techniques aim to capture overall patterns among each condition. 

\paragraph{Modeling Techniques}
In this section, we review the adopted ML models for detecting the cognitive status of individuals. We review deep learning models by categorizing the models as \textit{convolutional}-, \textit{recurrent}-, and \textit{attention}-based methods. Lastly, we review studies that integrated multiple models in their detection system. 

\textbf{Convolutional-based}: Meghanani~\etal~\cite{meghanani_recognition_2021} investigated a CNN layer applied to the word embedding and fasttext model that uses a bag of n-grams to capture the local word orderings. They fused the results from two different models by utilizing Bootstrap aggregation. Furthermore, the RNN model is integrated in~\cite{casanova_evaluating_2020} by applying CNN architecture to have sentence segmentation of the transcripts. 

\textbf{Recurrent-based}: the LSTM cells are a great method for extracting the sequential dependency of linguistic features. Another integration of LSTM is bidirectional-LSTM, which is explained in the name as it calculates the dependency of the sequential vectors in both directions of 1D data stream. Hong~\etal~\cite{hong_novel_2019} adapted two layers of bi-LSTM that take the word vectors after that the hidden 64 cells are passed into an attention layer to have the final context vector. Also, Roshanzamir~\etal~\cite{roshanzamir_transformer-based_2021} integrated an augmentation layer in their method while implementing a bi-LSTM to capture sequential information from the word embeddings.

\textbf{Attention-based}: the attention mechanism effectively extracts contextual representation from multiple vectors. Hong~\etal~\cite{hong_novel_2019} proposed different attention layers that are added together to have a final context vector. They used multi-weight attention to better capture semantic and grammatical features in each sentence. 

The Transformers are integrated to capture the self-attention from sequential input vectors. Tsai~\etal\cite{tsai_efficient_2021} implemented a transformer encoder with one layer of self-attention with four heads to classify a sequence of words from their word embeddings. They have initialized their model to classify the transcripts based on the cognitive conditions. Similarly, Fard~\etal~\cite{fard2024linguistic} have proposed a framework that utilized Transformer encoder modules to capture the sentence cross attention of participants' transcripts from sentence embeddings. They also proposed a loss function, InfoLoss, to enhance the cognitive detection of participants in their framework.  

\textbf{Multi-models}: studies have implemented several deep learning models with different architectures and have an overall decision of AD/ADRD prediction. Khan~\etal~\cite{khan_stacked_2022} have integrated three DL models in parallel (CNN, CNN followed by bi-LSTM, and bi-LSTM) then concatenate to a dense layer to make the final prediction. Another integration is conducted in~\cite{liu_improving_2022}, where two modules are trained on the same data. The first one is the G-Net Module to extract common features and the other one is the P-Net Module to purify the extracted features. These modules are transformer-based architecture. 

Chen~\etal~\cite{chen_attention-based_2019} implemented a 1D CNN layer followed by an attention layer to capture local features during Linguistic feature extraction, whereas the implementation of Bi-GRU to extract the global features of the transcripts before predicting the cognitive condition. They concatenated the local and global features before predicting the participants' conditions. Moreover, Wang~\etal~\cite{wang_explainable_2021} integrated the CNN layer and Attention layer to propose a C-attention network these models use a transformer encoder backbone network. This ensures the efficiency of extracted local features among part-of-speech (PoS) and sentence embedding features.

Language-based models consider the generated words or sentences of participants in detecting their cognitive status. The detection methods are achieved by transcribing the speech of participants, feature extraction (i.e., word embeddings), modeling integration, and classification/regression assigning. Using linguistic features, these methods with different approaches can distinguish the participants based on their cognitive conditions. Although there are attempts to detect cognitive impairment in various languages, it still requires more investigation on linguistic biomarkers across languages. For better organization, Table~\ref{table.linguistic_summary} illustrates the reviewed studies and contains the evaluation of these approaches.

\subsubsection{Acoustic and Linguistic Intergration}

\newcolumntype{C}[1]{>{\centering\arraybackslash}m{#1}}  

\begin{table*}[]
\centering
\caption{A summary of the reviewed studies that utilized acoustic and linguistic features}
\label{table.acoustic_linguistic_summary}
\rowcolors{2}{gray!10}{white}
\resizebox{\textwidth}{!}{
\begin{tabular}{cC{1cm}C{1cm}C{3.5cm}C{3.5cm}C{3cm}C{3cm}ccc}
\rowcolor{gray!30}
\hline
& & & \multicolumn{2}{c}{Features}                                                &   &             & \multicolumn{3}{c}{Evaluation (\%)} \\  \cline{4-5} \cline{8-10}
\rowcolor{gray!30}
\multirow{-2}{*}{Autors (Year)} & \multirow{-2}{*}{Dataset} & \multirow{-2}{*}{Language}      & Acoustic                                      & Linguistic                                                             &        \multirow{-2}{*}{Fusion}         & \multirow{-2}{*}{Classification Model}                  & ACC        & F1         & AUC  \\ \hline

Mittal~\etal (2021)~\cite{mittal_multi-modal_2021}                  &     &  & raw speech                                    & WE                                                         & Late fusion                                   &  acoustic (CNN),  linguistic (FastText-CNN, BERT, sentenceBERT)                  & 85.3       & 84.4       & 92.1 \\
Zolnoori~\etal (2023)~\cite{zolnoori_adscreen_2023}                &   \multirow{-2}{*}{Pitt}      &                           & MFCC, formant frequencies, voice intensity    &  semantic disfluency,  lexical diversity, syntactic, WE  & JMIM for feature selection                    & LSTM, CNN, traditional ML                                                      & 89.6       &            &      \\
Edwards~\etal (2020)~\cite{edwards_multiscale_2020}                 &   &                           & ComParE                                       & FastText, word2vec, Sent2Vec, StarSpace                                &                                               &                                                                                & 92.6       & 92.3       &      \\
Syed~\etal (2020)~\cite{syed_automated_2020}                   &                          &                           & speech paralinguistics, VGGish, ComParE, IS10 & WE                                                        &                                               & SVM                                                                            & 85.4       &            &      \\
Balagopalan~\etal (2021)~\cite{balagopalan_comparing_2021}             &                          &                           & lexico-syntactic feature            &                                                                        &                                               & SVM, RF, BERT                                                                  & 83.3       & 83.3       & 83.3 \\
Syed~\etal (2021)~\cite{syed_automated_2021}                    &                          &                           & IS10, ComParE, other paralinguistic           & syntactic, readability, lexical, WE                        & pooling function                              & SVM, LR                                                                        & 91.7       &            &      \\
Rohanian~\etal (2020)~\cite{rohanian_multi-modal_2020}                &                          &                           & prosodic, voice quality, spectral             & WE                                                 & Gated Layer                                   & Bi-LSTM                                                                        & 79.2       &            &      \\
Ilias and Askounis (2022)~\cite{ilias_multimodal_2022}      &                          &                           & log Mel-spectrograms                          &  WE                                                    &                                               & BERT, ViT, Co-Attention with shifting gate                                     & 90         &            &      \\
Li~\etal (2023)~\cite{li_leveraging_2023}                     &                          &                           & Whisper features, wav2vec2.0, wavLM   & BERT                                                                   & weighted sum or maximum layer, attention pool  & MLP                                                                            & 91.4       & 91.4       &      \\
Campbell~\etal (2020)~\cite{campbell_alzheimers_2020}                &                          &                           & i-vector, x-vector, rhythmic features         & WE                                                  & averaging scores                              & linguistic: RNN, acoustic: SVM                                                 & 82.4       & 81.9       & 90.2 \\
Cummins~\etal (2020)~\cite{cummins_comparison_2020}                &                          &                           & BoAW of MFCC, log-Mel, ComParE feature sets   & WE, Bi-LSTM: words and sentences                    & acoustic                                      & Bi-LSTM with attention                                                         & 85.2       & 85.4       &      \\
Mahajan and Baths (2021)~\cite{mahajan_acoustic_2021}      &          \multirow{-10}{*}{ADReSS}           &       \multirow{-12}{*}{English}        & raw speech                                    & GloVe and PoS                                                          & Dense layer                                   &  acoustic (dense layers + GRU),  linguistic (CNN + Bi-LSTM + attention)  & 72.9       &            &      \\
Pan~\etal (2021)~\cite{pan_using_2021}                    &   &   &                                               &                                                                        &                                               &                                                                                & 84.5       &            &      \\
Rohanian~\etal (2021)~\cite{rohanian_alzheimers_2021}               &                          &                           & feature set from CONVAREP                     &  GloVe after ASR, word probabilities, disfluencies, unfiled pauses     & Gating                                        & Bi-LSTM                                                                        & 84         &            &      \\
Pappagari~\etal (2021)~\cite{pappagari_automatic_2021}            &                          &                           & x-vector, VGGish, eGeMAPS                     & BERT embeddings                                                        &                                               &  Acoustic (ResNet-34),  Linguistic (fine-tune BERT)                             & 84.1       &            &      \\
Wang~\etal (2021)~\cite{wang_modular_2021}                  &        \multirow{-4}{*}{ADReSSo}            &   \multirow{-4}{*}{English}          & EMobase, IS10, VGGish, x-vector               & TTR, PoS                                                               &                                               & CNN and multi-head attention                                                   & 80.3       & 82.5       &      \\
\rowcolor{white}
\multicolumn{4}{l}{WE: word embedding.} & & & & & \\

\end{tabular}}
\end{table*}

In this section, we cover studies that utilized acoustic and linguistic features of individual speech to detect cognitive impairment. These studies proposed methods to integrate or investigate the feasibility of selected features in the detection systems. In general, merging two different types of features in deep learning models helps boost the performance~\cite{gao2020survey}.

\paragraph{Data preprocessing and feature extraction}

The preprocessing steps and features extraction methods have been introduced previously in Sec.\ref{subsubsec.acoustic.preprocessingAndFeatures} for acoustic methods and in Sec.\ref{subsubsec.lang.preprocessingAndFeatures} for linguistic methods.

\paragraph{Modeling Techniques}
The modeling techniques employed in these studies can be divided into traditional machine learning methods and deep learning methods, each offering unique benefits in processing and analyzing speech data for cognitive impairment detection.

\textbf{Traditional Machine Learning Methods}

Traditional machine learning models such as logistic regression (LR), SVM, and RF remain valuable due to their robustness on small datasets and ease of implementation. Studies by Syed~\etal~\cite{syed_automated_2020}, Edwards~\etal~\cite{edwards_multiscale_2020}, and Campbell~\etal~\cite{campbell_alzheimers_2020} have effectively utilized these models either standalone or as classification heads following feature extraction with deep learning backbones.

\textbf{Deep Learning Methods}

The rise of deep learning methodologies marks a significant advancement in speech data analysis. RNNs, particularly LSTM and GRU, are adept at handling sequential data. Rohanian~\etal~\cite{rohanian_multi-modal_2020} and Mahajan~\etal~\cite{mahajan_acoustic_2021} successfully implemented RNNs to process linguistic and acoustic data, respectively. Bi-directional LSTM variants further enhance performance by utilizing contextual information from both directions in a sequence~\cite{rohanian_alzheimers_2021}. 

CNNs are powerful for extracting crafted features through various convolutional and pooling layers, followed by fully connected layers. They leverage the hierarchical pattern extraction capabilities from structured data, highlighted in the research conducted by Koo~\etal~\cite{koo_exploiting_2020} and Mittal~\etal~\cite{mittal_multi-modal_2021}.

Transformer-based models like BERT and its variants (e.g., RoBERTa~\cite{liu2019roberta}, DistilBERT~\cite{sanh2019distilbert}) reflect the latest developments in contextual language understanding, as demonstrated by~\cite{ilias_multimodal_2022, liu_transfer_2022}. These models excel in capturing contextual nuances within linguistic data, capitalizing on attention mechanisms for improved abstraction and prediction accuracy. 

Hybrid models that combine acoustic and linguistic features via techniques like feature fusion, late fusion, and feature aggregation represent cutting-edge integrations in this research field. Feature fusion is critical in multi-modal learning to enhance model performance. Techniques such as concatenation, averaging, majority voting, attention mechanisms, and gating mechanisms are widely used. For example, Koo~\etal~\cite{koo_exploiting_2020} and Wang~\etal~\cite{wang_modular_2021} employed attention layers to weight and sum features from different modalities effectively. The gating mechanism, as presented in works by Rohanian~\etal~\cite{rohanian_multi-modal_2020} and Ilias~\etal~\cite{ilias_multimodal_2022}, similarly boosts the fusion of speech features. Late fusion involves collecting prediction scores from different branches and choosing the optimal result. Pappagari~\etal~\cite{pappagari_automatic_2021} demonstrated the efficacy of this technique by combining scores from multiple acoustic and linguistic models to determine the best overall prediction.

Mahajan and Baths~\cite{mahajan_acoustic_2021} and Wang~\etal~\cite{wang_modular_2021} exemplify these sophisticated systems, which enhance prediction performance by effectively synthesizing heterogeneous data types.

By leveraging and integrating these extracted features and modeling techniques, researchers can achieve robust and accurate detection of cognitive impairments from combined acoustic and linguistic speech data, as highlighted in the various reviewed studies. Thus, these studies underscore the progress and potential of speech-based screening tools for cognitive impairments, paving the way for impactful clinical applications. Table~\ref{table.acoustic_linguistic_summary} summarizes these studies and includes the evaluation of these studies, which is discussed later in the paper.

\subsection{Visual Modality} \label{subsec.Visual}

In this section, we review papers that utilize visual modals in detecting cognitively impaired participants by applying Deep Learning methods. These methods utilized different sources of datasets. Thus, this section will discuss the studies' data preprocessing and modeling techniques. The visual modality is mainly considering a facial or body video of participants. 

\subsubsection{Data Preprocessing and feature extraction}
The preprocessing and feature extraction of visual data depend on the type of the data. We discuss the different types of features that are sourced from a visual representation of participants. The video frames require feature extraction of most of the data causing the models to distinguish between cognitively impaired and normal cognitive participants. Feature extraction can be achieved directly by deep learning models or by using predefined features from the visual data. 

\textbf{Facial Attribute}: facial videos are utilized in several studies by targeting predefined facial features or facial measures. Auction units and facial expressions are examples of predefined facial features whereas facial landmarks and head poses are for facial measures. Also, in more specific facial features, eye gaze, eye blink rate, and lip movements are calculated from extracted facial landmarks to distinguish the participants regarding their cognitive status. Other studies have integrated holistic facial features of participants in their methodologies. 

The availability of pretrained deep learning models in extracting facial features has encouraged researchers to adopt these models during the detection of the cognitive status of participants. Although these models reached a state-of-the-art (SOTA) performance on public datasets, it should be important to mention that these datasets have an underrepresentation of older adults which is essential in features that differ with the subject's age.

Fei~\etal~\cite{fei_novel_2022} have proposed a framework from a modified version of MobileNetV2~\cite{sandler2018mobilenetv2} in extracting FER features from each frame of the videos. They have pretrained the model on a privately labeled elderly facial expression dataset, which is an advancement over other studies because it has fewer bias samples against elderly facial expressions. Similarly, Jiang~\etal~\cite{jiang2022automated} have extracted the FER of participants' faces while viewing a sequence of images.

Studies have calculated temporal facial features by tracking facial landmarks during the face recording of participants. This tracking result measures a useful aspect of facial representation of cognitive status. Alzahrani~\etal~\cite{alzahrani2021eye} calculated an eye blink rate using six facial landmarks for each eye to distinguish participants' cognitive status. Moreover, Tanaka~\etal~\cite{tanaka2019detecting} used facial landmarks around the participants' lips to segment their speech responses.

Zheng~\etal~\cite{zheng_detecting_2023} applied pre-trained models to extract face mesh using MediaPipe~\cite{lugaresi2019mediapipe}, an open-source library that provides trained DL models for various computer vision tasks. They also extracted Histogram of Oriented Gradients (HOG) features from video frames using OpenFace Library~\cite{baltrusaitis2018openface} involving detecting facial landmarks as well. They feature engineered the features from the video frames. They also extracted Action Units (AU) intensities from the HOG and then calculated the mean and variances of the AU intensities.

Other studies have integrated non-defined facial features in implementing deep learning models. These methods follow either embedding facial features or end-to-end detection frameworks. Firstly, embedded features can be extracted by utilizing an unsupervised learning approach to the data. Alsuhaibani~\etal~\cite{alsuhaibani2024mild}, for example, implemented a convolutional autoencoder to extract embedding facial features of participants from video frames. Secondly, the direct integration of data into the deep learning detection model would learn general facial representation during the detection task. Sun~\etal~\cite{sun_mc-vivit_2024} detection method involves a direct projection of spatial-temporal features into the proposed MC-ViViT model. 

\textbf{Gait Pattern}: the body's behavioral observations are collected with either video-based systems or sensor-based systems. In this section, we review gaits captured from the camera system whereas we review later in Sec \ref{subsec.Sensor} body features that are captured using sensors. These observations are intended to find abnormalities in walking or hand swing, and human gait in general. Studies have extracted the body skeleton points from depth cameras, commercially known as Kinect. After extracting the features, the body skeleton points are presented in a vector at each time instant.  

You~\etal~\cite{you_alzheimers_2020} identified key points from gait and EEG data, with the EEG data being downsampled from 5000Hz to 250Hz for improved processing speed. During the data preprocessing stage, coordinate transformation was applied to the collected gait data. You~\etal~\cite{you_alzheimers_2021} analyzed gait patterns by extracting 25 joint points, with Special features of gait including average speed, half-of-gait cycle and variation, stride length and variation, hand swing, and head posture variation. Aoki~\etal~\cite{aoki_early_2019} employed the Hilbert-Huang transform, to analyze time series data, in the preprocessing of sensor data. The collected sensor data were then segmented into sequences of coordinates representing body joints.

\textbf{Gaze Tracking:} Zuo~\etal~\cite{zuo_deep_2023} utilized visual attention heatmaps generated during a 3D visual task, using a non-invasive eye-tracking system. The patterns captured in these heatmaps function as the fundamental features used to train a model.

\subsubsection{Modeling Techniques} 
In this section, we review the methods adopted by studies that utilized visual modalities in detecting cognitive impairment. We first start with traditional machine learning models and then deep learning models. 

Since models are limited to certain dimensions, studies have integrated splitting techniques for training the models and merging decision-making for participants' cognitive condition prediction. Splitting a full video into segments before training the deep learning model is the method implemented in~\cite{zheng_detecting_2023,sun_mc-vivit_2024,alsuhaibani2024mild}. No window size is proven to be the best in detecting cognitive impairments from facial features. The window sizes of the data affect the DL models since the models' weights are updated based on the training batches. The global decision in detecting participants’ cognitive impairments occurs after merging the window labels. The majority voting method is integrated into making a decision more often than averaging the confidence of all segments~\cite{sun_mc-vivit_2024,zheng_detecting_2023,alsuhaibani2024mild}. 

This raises an issue of an imbalance of the number of segments within a participant's data or from the classes. works integrated different methods to overcome this issue. the imbalance issue of the number of samples from subjects with different cognitive conditions is considered in~\cite{sun_mc-vivit_2024,alsuhaibani2024mild}. Sun~\etal~\cite{sun_mc-vivit_2024} proposed a loss function to overcome the inter- and intra-class imbalance in the dataset. The proposed loss function moderated during their model training was able to boost the model toward more balance of the dataset. On the other hand,~\cite{alsuhaibani2024mild} integrated the weighted cross entropy loss function while training the model. Fei~\etal~\cite{fei_novel_2022} selected periods of the emotional occurrence of each participant manually which presented the most intense facial expression during the video. They integrated a traditional ML model (i.e., SVM) to classify the participants' cognitive conditions. Zheng~\etal~\cite{zheng_detecting_2023} explored two different sizes of segmentation within a video to be presented, namely 1024 or 512 frames within a segment. Introducing the segmentation increased the number of instances because they considered the mean and variances of AU intensities. Note that the PROMPT dataset~\cite{kishimoto_project_2020} contains videos with a frame rate of 30 fps. 

Zuo~\etal~\cite{zuo_deep_2023} focused on gaze tracking, utilizing a multi-layered comparison convolutional neural network (MC-CNN) for classification between individuals with AD and NC, utilizing similarity between pairs of heatmaps associated with AD and normal cognitive states.

These attempts to detect cognitive impairment using visual modalities and deep machine learning approaches are, however, still limited compared to other modalities such as speech and language. Although it has a late start compared to language, it gained momentum in recent years with the advancement of computational powers. Similar to other modalities, temporal information must be considered; however, visual modalities have a more dimensional representation. Thus, extracting features such as AU, FER, and skeleton joint points are widely adopted in the visual aspect. Consequently, end-to-end detection frameworks are rarely implemented. This is because they increase the model complexity dramatically. Table~\ref{table.visual_sensor_summary} shows a summary of these studies along with the evaluation of their approach.

\newcolumntype{C}[1]{>{\centering\arraybackslash}m{#1}} 

\begin{table*}[]
\centering
\caption{A summary of the reviewed studies that utilized visual and motoric mobility modalities}
\label{table.visual_sensor_summary}
\rowcolors{2}{gray!10}{white}
\resizebox{\textwidth}{!}{
\begin{tabular}{cccC{5cm}ccccc}
\hline
\rowcolor{gray!30}
 &  &     &       &  & \multicolumn{3}{c}{Evaluation (\%)} \\ \cline{6-8}
\rowcolor{gray!30}
\multirow{-2}{*}{Autors} & \multirow{-2}{*}{Dataset}  & \multirow{-2}{*}{Modality}    & \multirow{-2}{*}{Features}      & \multirow{-2}{*}{Classification Model}   & ACC      & F1       & AUC      \\ \hline

Sun~\etal (2024)~\cite{sun_mc-vivit_2024}              &           &   \multirow{4}{*}{Facial}                           & raw sequence of frames                                                                                 & ViViT with multi-branch classifier                            & 90.6     & 93       & 60.4     \\
Alsuhaibani~\etal (2024)~\cite{alsuhaibani2024mild}      & \multirow{-2}{*}{I-CONECT} &                              & AutoEncoder for facial features, interation features                                                   & Transformer encoder                                            & 87.5     & 89       & 87       \\
Zheng~\etal (2023)~\cite{zheng_detecting_2023}            & PROMPT                    &                              & Face Mesh                                                                                              & LSTM                                                          & 79       & 81       &   -       \\
Fei~\etal (2022)~\cite{fei_novel_2022}            &      &       & emotion occurance                                                                                      & SVM                                                            & 73.3     &     -     &      -    \\
Zuo~\etal (2023)~\cite{zuo_deep_2023}             &  & Eye gaze                     & heatmaps from eye gaze                                                                                 & multi-layed comparison CNN                                     & 83       & 81       &   -       \\
You~\etal (2020)~\cite{you_alzheimers_2020}            &                           & Gait and EEG                 & AST-GCN to extract features from gait, ST-CNN to extract features from EEG                             &                                                               &   93.1      &  -        & -         \\
You~\etal(2021)~\cite{you_alzheimers_2021}             &                           &                              & body movement statical measurement                                                                     & FC, LSTM, MLP                                                 & 90.5     &    -      &   -       \\
Ghoraani~\etal (2021)~\cite{ghoraani_detection_2021}        &                           &                              & PKMAS                                                     &                                           & 86       & 88       &     -     \\
Shahzad~\etal (2022)~\cite{shahzad_automated_2022}         &                           &                              &                                                                                                        &          \multirow{-2}{*}{SVM}                           & 70       &     -     &   -       \\
Bringas~\etal (2020)~\cite{bringas_alzheimers_2020}         &                           & daily activity               &  sequence of accelerater data                                                                    & CNN                                                          & 90.9     & 89.7     &   -       \\
El-Yacoubi~\etal (2019)~\cite{el-yacoubi_aging_2019}      &                           &  & horizontal and vertical velocities, acceleration, and jerk, direction and curvature, pen-lift duration & Bayes' classifer                                               & 74.3     &    -      &   -       \\
Cilia~\etal (2021)~\cite{cilia_handwriting-based_2021}           &                           &  \multirow{-3}{*}{{Handwriting}}  & peak vertical velocity and acceleration, jerk, pen pressure, strikes, age, education                    & SVM, DT, NN                                                    & 90.4     &   -       &   -       \\
Narasimhan~\etal (2020)~\cite{narasimhan_early_2020}      &       \multirow{-14}{*}{\rotatebox{90}{private}}                    & daily activity               &      sleep duration, cooking time, and walking speed            & LSTM                                            &    77.5      &  -        &    -     \\ \hline
\end{tabular}}
\end{table*}

\subsection{Other Data Modalities} \label{subsec.Sensor}
In this section, we review research that utilized data modalities other than speech, and videos. The data modalities include those acquired by wearable and non-wearable measuring devices to detect cognitive impairment or AD/ADRD. These data mainly capture the motoric mobility of subjects. We discuss the essential aspects such as the activities under consideration, the feature extraction process, and the methodologies implemented in each case.

\subsubsection{Data Preprocessing and Feature Extraction} \label{subsec.Sensor.Feature_Extraction}
In this section, we describe the process of extracting features to maintain a vital role in the automated pipeline for detecting cognitive impairment. This primary process, dependent on the type of data employed, is fundamental for identifying patterns indicative of cognitive conditions.

\textbf{Gait Data:} Key points that are extracted from gait data, are the main features utilized for detecting cognitive impairment. 
Bringas~\etal~\cite{bringas_alzheimers_2020} recorded acceleration changes in the X, Y, and Z axes over time, along the temporal dimension, to predict the AD stage. Ghoraani~\etal~\cite{ghoraani_detection_2021} used the PKMAS to extract gait features from the Zenomat system and GAITRite software for the GAITRite system. Shahzad~\etal~\cite{shahzad_automated_2022} collected specifically mobility data, including a combination of a triaxial accelerometer and triaxial gyroscope.

\textbf{Handwriting:} Handwriting-related features are utilized in the study of cognitive impairment assessments. El-Yacoubi~\etal~\cite{el-yacoubi_aging_2019} analyzed text-based features, including horizontal and vertical velocities, the first derivative representing acceleration, second derivatives indicating jerk, direction and curvature, and the duration of pen-lifts. Cilia~\etal~\cite{cilia_handwriting-based_2021} utilized handwriting-related features, including peak vertical velocity and acceleration, average absolute velocity, normalized jerk, and pen pressure, alongside the analysis of other features such as the number of strokes, age, and education.

\textbf{Daily Routines:} Narasimhan~\etal~\cite{narasimhan_early_2020} identified a set of 12 features associated with diverse daily activities such as mobility, morning care, personal hygiene/grooming, eating, and memory, alongside health assessment score observed at the assessment time point, containing both physical and cognitive/behavioral aspects.

\subsubsection{Modeling Techniques}\label{subsec.Sensor.Proposed_Methodologies}
In Sec.~\ref{subsec.dataset.sensor-based} and Sec.~\ref{subsec.Sensor.Feature_Extraction} we described the data source and the feature extraction processes, respectively. In this section, we review the prediction methodology of the research that utilized movement measured data for the detection of cognitive impairment. From a high-level perspective, the following section studies two main categories: traditional machine learning methods and deep learning methods.

\textbf{Traditional Machine Learning Methods:}
In the following, we review the research that utilized traditional machine learning algorithms including SVM, Decision Tree (DT), and Bayesian networks to predict cognitive impairment.

Ghoraani~\etal~\cite{ghoraani_detection_2021} focused on gait data, utilizing an SVM to distinguish between MCI and AD based on gait data and MoCA scores separately. The analysis utilized a three-channel SVM model, and the ultimate prediction was performed via majority voting. Shahzad~\etal~\cite{shahzad_automated_2022} focused on gait data. The proposed approach involved combining sensor data with a cognitive task, specifically a 10-meter walkway with a subject wearing inertial sensors consisting of mobility data gathered using a triaxial accelerometer and triaxial gyroscope. The analysis utilized a multi-kernel SVM to distinguish between individuals with MCI and NC. El-Yacoubi~\etal~\cite{el-yacoubi_aging_2019} proposed a Bayesian classifier aiming to detect AD based on handwriting patterns. Cilia~\etal~\cite{cilia_handwriting-based_2021} conducted a handwriting study and proposed an ensembled classifier that included DT, Neural Networks (NN), and SVM to predict cognitive impairments

\textbf{Deep Learning Methods:}
In the following, we review the research that utilized deep learning models to predict cognitive impairment. 

Narasimhan~\etal~\cite{narasimhan_early_2020} focused on daily routines, combining activities tracking data with health assessment records as input to an LSTM model to predict the stage of AD. You~\etal~\cite{you_alzheimers_2020} conducted a gait data analysis and proposed a deep model including attention-based spatial-temporal graph convolutional networks (AST-GCN) as well as a spatial-temporal convolutional network (ST-CNN) to distinguish between HC and MCI or AD.

Bringas~\etal~\cite{bringas_alzheimers_2020} focused on gait data and proposed an MLP where the input for the network was a 10804 $\times$ 3 tensor, corresponding to the time dimension and axes (x, y, z), to predict and specify the stage of AD. You~\etal~\cite{you_alzheimers_2021} focused on gait data and proposed 1-dimensional CNN, where the input is 1 hour of accelerometer data and the output is the probability scores concerning MCI and AD.

These studies have attempted to detect cognitive impairment using data collected through measuring devices. Meanwhile, Table~\ref{table.visual_sensor_summary} summarizes these studies, as discussed in this section, while also mentioning the evaluation results.

\section{Performance Evaluation} \label{sec.Performance}

In this section, we discuss the performance of the methods reviewed in Section \ref{sec.Modeling and Modalities} through common evaluation metrics across all the studies. We will further explore these methods and analyze the justifications for their results. 

Researchers use various methods to evaluate their models, typically through test subsets within the dataset or by applying statistical evaluation methods. Cross-validation, a common statistical evaluation method, is used to evaluate models for detecting cognitive conditions, with 5-fold, 10-fold, and leave-one-out being the most frequently employed methods. Some studies, such as Bertini~\etal~\cite{bertini_automatic_2022}, empirically selected 20-fold cross-validation. The number of folds considers the data bias, variance, and computational complexity. Ideally, it depends on the dataset size; thus, the number of folds should be increased when the dataset size is small. On the other hand, fewer folds are sufficient for large datasets. Typically, 5 to 10 folds are standard in machine learning.

The evaluation metrics in studies using DL models include accuracy, area under the receiver operating characteristic (ROC) curve (AUC), F-score, precision, recall, sensitivity, and specificity, while classification error rate and ROC are rarely used. The root mean squared error (RMSE) is the only metric used for regression tasks. Most classification tasks focus on binary detection, with accuracy being widely reported despite its limitations on imbalanced datasets. Therefore, accuracy is often supplemented by other metrics that account for false positives (FP) and false negatives (FN), with the F1 score being the second most commonly used metric among the reviewed studies.

The evaluation of methods and approaches for cognitive impairment detection can be better understood by considering the various modalities and distinct performance metrics used across different studies. In the following sections, we discuss detection performance categorized by their modalities by briefly mentioning the approach and the evaluation results.

\subsection{Acoustic Modality}

Syed~\etal~\cite{syed_static_2020} employed Bi-LSTM with IS10-paralinguistics feature sets on the ADReSS dataset, achieving a classification accuracy of 74.55\%. This performance highlights the potential of traditional feature sets in conjunction with deep learning models, specifically the importance of paralinguistic features in detecting cognitive impairments. Chlasta and Wolk~\cite{chlasta_towards_2021} utilized the VGGish model to extract features with a DemCNN model on the same dataset, resulting in a lower accuracy of 63.6\%. However, their precision, recall, and specificity metrics all scored 69.2\%, indicating a balanced performance across these metrics. This suggests that while the overall accuracy was lower, the model maintained consistent detection capabilities for both classes. Meghanani~\etal~\cite{meghanani_exploration_2021} achieved a classification accuracy of 64.58\% with a CNN-LSTM model using MFCCs, and their derivatives. Including spectrogram-based features and temporal dynamics through LSTM layers improves the performance compared to the DemCNN detecting model using features extracted from VGGish.

Gauder~\etal~\cite{gauder_alzheimer_2021} reported an accuracy of 78.87\% on the ADReSSo dataset using a multi-feature approach that included eGeMAPS, Trill, Allosaurus, and Wav2vec2.0 features processed through a series of 1D CNNs. This indicates that a multi-faceted approach to extract comprehensive acoustic features can substantially enhance the model’s performance in detecting cognitive impairments.

Several studies using the Pitt dataset have shown varying levels of success. Liu~\etal~\cite{liu_detecting_2021} used spectral features of phonemes with a CNN-Bi-LSTM-attention pool-FC layer and achieved an accuracy of 82.59\%, with a recall of 85.24\% and a precision of 82.94\%, suggesting their model was effective in identifying subtle variations in phonetic features related to cognitive impairments. Furthermore, Kumar~\etal~\cite{kumar_dementia_2022} reported an accuracy of 87.6\% by employing a RF classifier with an extensive set of 44 acoustic features while the PRCNN achieved an accuracy of 85\% and an F1-score of 85.1\% using 62 features within 25ms segments. Similarly, Warnita~\etal~\cite{warnita_detecting_2018} reported an accuracy of 73.6\% using GCNN utilizing sets of Interspeech features. Additionally, Bertini~\etal~\cite{bertini_automatic_2022} achieved a classification accuracy of 90.7\% and an F1-score of 88.5\% using an Autoencoder with GRU encoder and MLP to classify log-Mel spectrogram images, emphasizing the value of advanced feature extraction and autoencoder methodologies in enhancing classification performance. Similarly, Pranav~\etal~\cite{pranav_early_2023} employed a ViT on log-Mel spectrograms, attaining an accuracy of 85.7\% and an F1-score of 92.3\%.

Utilizing non-English datasets, Bertini~\etal~\cite{bertini_automatic_2021} have implemented an Autoencoder with GRU encoder and MLP to classify log-Mel spectrogram features on a private Italian dataset with a detection accuracy of 90.6\% and an F1 of 90.7\%. Moreover, Nishikawa~\etal~\cite{nishikawa_detecting_2021} reported a classification accuracy of 90.8\% with a 1-d CNN-LSTM model on a private Japanese dataset. In addition, using another approach, Nishikawa~\etal~\cite{nishikawa_machine_2022} achieved an accuracy of 89.4\% using ViT\_b16, demonstrating consistency and robustness across different languages and feature sets. On the other hand, Rodrigues Makiuchi~\etal~\cite{rodrigues_makiuchi_speech_2021} have utilized the PROMPT dataset~\cite{kishimoto_project_2020} using GCNN to achieve a detection accuracy of 80.8\% among Japanese individuals.

Overall, studies that extracted comprehensive acoustic features achieved better performance compared to studies that used a few types of acoustic features. Although studies that utilized acoustic feature sets achieved an adequate detecting performance, they are on the lower scale of detecting cognitive conditions. Log Mel-spectrogram contains an indication of the speaker's cognitive status. Thus, advanced computer vision models reached a high detection performance. Table.~\ref{table.acoustic_summary} shows the studies results of acoustic feature approaches. 

\subsection{Linguistic Modality}

In utilizing the ADReSS dataset, Searle~\etal~\cite{searle_comparing_2020} implemented an SVM classifier on features on DistilBERT word embeddings and TF-IDF and achieved an accuracy of 81\%. This study combines statistical linguistic analysis with modern embeddings for cognitive impairment detection. Additionally, the integration of LASSO regression yielded an RMSE of 4.58, demonstrating substantial precision in regression tasks. Meghanani~\etal~\cite{meghanani_recognition_2021} also utilized word embeddings in conjunction with a CNN, highlighting the potential of convolutional networks to capture spatial patterns within linguistic features, reaching an accuracy performance of 83.3\%.  

Moreover, Yuan~\etal~\cite{yuan_pauses_2021} achieved an accuracy of 89.6\% by fine-tuning LLMs like BERT and ERNIE, which encoded participants' pauses into transcripts. This suggests that leveraging large pre-trained models and fine-tuning them for domain-specific tasks significantly enhances detection capabilities. Liu~\etal~\cite{liu_transfer_2022} also reported an accuracy of 88\% by fine-tuning DistilBERT with logistic regression on the same dataset, reinforcing the efficacy of using compact and powerful Transformer models in cognitive impairment classification. Saltz~\etal~\cite{saltz_dementia_2021} utilized multiple word embeddings (BERT, XLNet, ELECTRA) and type-token-ratio, and obtained varied results across different datasets (ADReSS, Pitt, UW). Specifically, they achieved accuracies of 76\%, 90\%, and 74\% on the Pitt dataset, augmented ADReSS, and UW, respectively, highlighting the robustness of their proposed methods across different datasets. Guo~\etal~\cite{guo_crossing_2021} achieved an accuracy of 97.9\% using fine-tuned BERT, aided by integrating WLS controls with the ADReSS dataset. The AUC of 99.2\% further demonstrates this model's exceptional discriminative ability due to supplementary training samples.

For the Pitt dataset, several studies highlighted varying successes. Fritsch~\etal~\cite{fritsch_automatic_2019} achieved an accuracy of 85.6\% using an LSTM model, while Chen~\etal~\cite{chen_attention-based_2019} reported an accuracy of 97.42\% by utilizing a combination of BiGRU and CNN models to extract global and local contextual features via GLoVe embeddings. This indicates the potential of sophisticated architectures to capture nuanced linguistic features. Roshanzamir~\etal~\cite{roshanzamir_transformer-based_2021} implemented an augmentation layer with Bi-LSTM or logistic regression and achieved an accuracy of 88.08\%, with precision and recall scores of 87.23\% and 90.57\%, respectively. This study underscores the importance of augmentation in improving model robustness and performance. Liu~\etal~\cite{liu_improving_2022} and Tsai~\etal~\cite{tsai_efficient_2021} also achieved accuracies of 93.5\% and 84\%, respectively, using transformer encoders, reaffirming the versatility and effectiveness of Transformer architectures on cognitive impairment detection.

Additionally, Khan~\etal~\cite{khan_stacked_2022} applied a stacked DNN combining CNN, Bi-LSTM, and MLP, achieving 93.31\%, 92.2\%, and 85.7\% for accuracy F1, and AUC, respectively, indicating that hybrid models leveraging both local and sequential features are highly effective. Wen~\etal\cite{wen_revealing_2023} further verified the efficacy of combining syntactic features with attention mechanisms and CNNs, achieving an accuracy of 92.2\%. Wang~\etal~\cite{wang_explainable_2021} utilized POS tags and sentence embeddings in a C-attention model, achieving an accuracy of 91.5\% and an AUC of 97.7\%, emphasizing the effectiveness of contextual and syntactic features in cognitive impairment detection. Similarly, Fard~\etal~\cite{fard2024linguistic} have generated sentence embeddings and utilized a Transformer encoder with proposed infoLoss for calculating the cost function to achieve an accuracy of 85.16\% and an AUC of 84.75\% on the I-CONECT dataset. 

In studies involving multiple datasets, Alkenani~\etal~\cite{alkenani_predicting_2021} reported an AUC of 98.1\% and 99.47\% on spoken and writing datasets, respectively, using lexicosyntactic and n-gram features in stacked fusion models, demonstrating the efficacy of ensemble learning across different linguistic contexts. Finally, studies on non-English datasets showed promising results with Casanova~\etal~\cite{casanova_evaluating_2020} achieving 75\% accuracy using RNNs for Portuguese data, and AI-Atroshi~\etal~\cite{ai-atroshi_automated_2022} achieving accuracies of 90.28\% and 86.76\% on Hungarian data using MLP with Gaussian mixture model and deep belief network.

Overall, we can conclude that large language models have the lead in detection performance. Nevertheless, other attempts using various combinations of features and classification models also demonstrated impressive detection performance. In addition, fine-tuning LLMs with control participants from other datasets helped the study stand out in the evaluation results. Table.~\ref{table.linguistic_summary} shows the overall approaches of linguistic features and their results.

\subsection{Acoustic and Linguistic Modalities}

In studies utilizing the ADReSS dataset, Campbell~\etal~\cite{campbell_alzheimers_2020} employed RNN for linguistic features and SVM for acoustic features, achieving an accuracy of 82.41\% using an averaging score fusion approach. Cummins~\etal~\cite{cummins_comparison_2020} improved these results, obtaining an 85.2\% accuracy with Bi-LSTM models, leveraging attention mechanisms tailored for different acoustic features. Edwards~\etal~\cite{edwards_multiscale_2020} reached an accuracy of 92.6\%, underscoring the effectiveness of linguistic models using embeddings like FastText.

Koo~\etal~\cite{koo_exploiting_2020} experimented with CNN and Bi-LSTM models but evaluated them differently, achieving a comparative baseline increment, whereas Rohanian~\etal~\cite{rohanian_multi-modal_2020} achieved an accuracy of 79.2\% with Bi-LSTM models using gated layer aggregation, both for classification and regression tasks. Syed~\etal~\cite{syed_automated_2020} highlighted the strengths of traditional SVM models, achieving 85.42\% accuracy, with regression RMSE of 4.3.

Later studies further diversified approaches. Balagopalan~\etal~\cite{balagopalan_comparing_2021} successfully showed BERT's ability in classification with an accuracy of 83.32\% while employing linear and ridge regression with an RMSE of 4.56. Meanwhile, Mahajan and Baths~\cite{mahajan_acoustic_2021} resulted in an accuracy of 72.92\% after applying dense layer fusion. Syed~\etal~\cite{syed_automated_2021} detecting cognitive impairment with accuracy reaching 91.67\% using an SVM model, emphasizing the efficacy of linguistic features.

The research by Ilias and Askounis~\cite{ilias_multimodal_2022} used ViT and co-attention gates, achieving an RMSE of 3.61, while Li~\etal~\cite{li_leveraging_2023} combined Whisper, BERT, and task-correlated features to attain an accuracy of 91.41\% with a solid AUC of 91.38\%. Moreover, Rohanian~\etal~\cite{rohanian_alzheimers_2021} used Bi-LSTM with gating and disfluency features, achieving an accuracy of 84\%, and Pappagari~\etal~\cite{pappagari_automatic_2021} combined ResNet-34 for acoustic and fine-tuned BERT models for linguistics, highlighting superior linguistic performance and achieving RMSE scores of 3.85.

Studies by Mittal~\etal~\cite{mittal_multi-modal_2021} and Zolnoori~\etal~\cite{zolnoori_adscreen_2023} on the Pitt dataset demonstrated deep models with late fusion, yielding accuracy figures of 85.3\% and 89.55\%, respectively, the latter leveraging JMIM selection techniques.

Overall, these studies collectively demonstrate that optimal performance in cognitive impairment detection is attained through deep learning architectures that effectively fuse acoustic-linguistic features, where advanced word embeddings like BERT frequently act as significant enhancers. The outcomes suggest room for continued exploration into model architectures and feature fusion methodologies to further refine and personalize cognitive health assessments. Table.~\ref{table.acoustic_linguistic_summary} presents the reviewed studies of their approaches and evaluation results.

\subsection{Visual Modality}

Several studies have concentrated on using facial features to detect cognitive impairment, leveraging the advancement of deep-learning techniques. Zheng~\etal~\cite{zheng_detecting_2023} utilized face mesh, Histograms of Oriented Gradients, and Action Units extracted from facial images, achieving an accuracy of 79\% with an LSTM for face mesh and HOG. Similarly, Alsuhaibani~\etal~\cite{alsuhaibani2024mild} applied transformer encoders on latent facial and interaction features from the I-CONECT dataset, reaching an accuracy of 87.5\%. Using autoencoders likely facilitated the extraction of high-level abstract features, which, when combined with the transformer encoder models, led to enhanced predictive capabilities. The F1-score of 89\% underscores the model's reliability in distinguishing cognitive impairment. Continuing with the I-CONECT dataset, Sun~\etal~\cite{sun_mc-vivit_2024} demonstrated the efficacy of a ViViT model with a multi-branch classifier applied to sequences of video frames, achieving an accuracy of 90.63\%. However, this study reported an AUC of 60.42\%. On the other hand, Fei~\etal~\cite{fei_novel_2022} employed MobileNetV2 for facial expression recognition and emotion occurrence, paired with an SVM, yielding an accuracy of 73.3\%. This indicates that while deep CNNs like MobileNetV2 can effectively extract facial expressions, additional advanced feature processing might be necessary to improve cognitive impairment classification outcomes.

Gait analysis has also been a key focus on visual modalities. Aoki~\etal~\cite{aoki_early_2019} utilized the Hilbert-Huang Transform for gait features from Kinect captured videos, classified using an SVM classifier, achieving an AUC of 74.7\%. This transformation likely enhanced the interpretability of non-linear and non-stationary gait patterns associated with cognitive changes. You~\etal~\cite{you_alzheimers_2020} integrated features from gait and EEG data using AST-GCN for gait and ST-CNN for EEG, yielding a classification accuracy of 93.09\% for HC vs MCI. The excellent performance suggests that combining multiple modalities can significantly improve the granularity and accuracy of cognitive impairment detection. You~\etal~\cite{you_alzheimers_2021} investigated gait features by rearranging data points and analyzing average speed, stride length, and other gait cycle variations, using FC, LSTM, and MLP models. This study reported an accuracy of 90.48\%, with a recall of 92\% and a specificity of 88.24\%.

From eye gaze data, Zuo~\etal~\cite{zuo_deep_2023} created heatmaps of individuals, implementing a multi-layer comparison CNN. With an accuracy of 83\%, the study highlights eye gaze as a significant predictor, endorsed by an F1 score of 81\%, offering valuable insights into eye gaze association with cognitive status.

Overall, studies utilizing visual modalities have shown promise in adopting this modality for cognitive impairment detection. However, further studies are required to ensure the robustness of these studies. Table.~\ref{table.visual_sensor_summary} shows the conducted studies methods and their evaluation results. 

\subsection{Other Data Modalities}
In handwriting analysis, El-Yacoubi~\etal~\cite{el-yacoubi_aging_2019} extracted features such as horizontal and vertical velocities, acceleration, and jerk among others. Utilizing a Bayes classifier, the study achieved an accuracy of 74.3\%, highlighting the critical role of velocity-based features in enhancing detection performance. Moreover, Cilia~\etal~\cite{cilia_handwriting-based_2021} explored peak vertical velocity and pen pressure, employing a combination of SVM, DT, and neural networks. Achieving a 90.4\% accuracy, the study demonstrated the potential of integrating multiple classifiers for nuanced cognitive assessment, especially distinguishing patient handwriting from healthy individuals using SVM.

By capturing gaits from motion measuring devices, Ghoraani~\etal~\cite{ghoraani_detection_2021} instrumented gait analysis with PKMAS for feature extraction alongside statistical feature selection methods. The SVM classifier achieved an accuracy of 86\% in distinguishing healthy subjects from those with MCI and AD. Shahzad~\etal~\cite{shahzad_automated_2022} also investigated gaits features by utilizing a multi-kernel SVM, resulting in an accuracy of 70\%.

For daily activities, Bringas~\etal~\cite{bringas_alzheimers_2020} adopted a CNN to analyze accelerometer data from daily activities. These features were subsequently classified using an MLP, resulting in an accuracy of 90.91\% with an F1 score of 89.7\%. This emphasizes the model's robust classification capabilities. 

Ultimately, motoric mobility ability is an indicator of cognitive conditions by using deep learning methods. However, it lacks more studies across various measuring devices. Table.~\ref{table.visual_sensor_summary} shows the reviewed studies along with their evaluation results.

\section{Challenges and Future Directions} \label{sec.Challenges}

In Sec.~\ref{sec.Modeling and Modalities} \& \ref{sec.Performance}, we reviewed and explained the growth and success of deep machine learning methods using non-invasive indicators including speech, vision, and movement-measured data for the detection of cognitive impairments, such as AD/ADRD. Despite their success, several challenges still exist that should be addressed to realize the full potential of non-invasive indicators. This section categorizes these challenges and suggests future research directions to advance the field.

\subsection{Challenges}
We classify the existing challenges into three main categories: \textit{data-related}, \textit{methodological}, and \textit{medical adaptation}. Each category presents unique obstacles that need to be dealt with to improve the efficacy and reliability of ML-based cognitive impairment detection methods.

\subsubsection{Data-Related Challenges}
Despite the success of deep learning models in various domains~\cite{fard2022ad, fard2022sagittal, fard2022facial, fard2021asmnet}, in general, these models have a heavy reliance on data. Specifically, for medical applications, the dependency on vast amounts of high-quality labeled and unlabeled data poses several challenges including the standardization, diversity, and data accessibility used for training and validating deep learning models. In the following, we delve into these data-related challenges.

\textbf{Standardization:} Medical researchers often capture various indicators of cognitive decline. Next, these indicators are examined by technical researchers with the ultimate goal of utilizing and integrating them into deep machine-learning methods for developing AD/MCI detection systems. This process needs to be standardized. Specifically, standardizing data collection is crucial as it reduces preprocessing requirements and ensures consistency~\cite{lin2024data}. This standardization improves data quality, making preprocessing more efficient and reliable. For instance, gaits patterns are captured using either camera-based systems~\cite{aoki_early_2019} or motion sensing-based systems~\cite{shahzad_automated_2022}. Although they extracted different feature maps, the movement indicators should align with the neuroscientific explanation for abnormalities due to cognitive decline. 

\textbf{Diversity:} Deep learning models often face challenges due to data imbalance, where the majority class dominates the dataset. This issue can lead to biased models performing well in the majority class. Thus, it makes the sensitivity or the specificity values skewed~\cite{johnson2019survey}. Despite various strategies being used to cope with this issue, such as data augmentation and re-sampling techniques~\cite{mollahosseini2017affectnet}, and algorithmic methods~\cite{f2023alyzer, fard2022ad}, a balanced dataset leads to investigating the data features rather than resolving the issues. Addressing this issue is crucial for developing reliable detection methods. 

Most public datasets (see Sec.~\ref{sec.Datasets} for more details) consider a balanced representation of different conditions and genders of individuals. However, the diverse demographic representation of race and ethnicity is either limited as it is mentioned in the I-CONECT dataset~\cite{dodge2024internet} or not acknowledged~\cite{becker1994natural, kishimoto_project_2020}.

\textbf{Data Accessibility:} The data accessibility strengthens the development of deep learning models in general. As we discussed in Sec.~\ref{sec.Datasets}, some modalities do not have available public datasets to researchers. Thus, this is essential in configuring and comparing the developed DL-based systems. Due to several reasons, such as privacy concerns, intellectual property rights, or institutional policies, researchers across various laboratories would not be able to gain access to restricted or proprietary datasets.

This lack of access to data creates barriers to entry for many research groups, particularly those with limited resources, as they are unable to leverage valuable data that could drive significant advancements in their work. This limitation not only restrains innovation but also hinders the collaborative nature of scientific research, where sharing data and results can lead to more rapid and meaningful developments.

Generalization and bias are critical considerations in studies that aim to be utilized as screening tools. Generalization refers to the ability of a model to perform well on new, unseen data, outside of the original study sample. This is crucial in cognitive impairment detection to ensure that findings can be reliably applied to diverse populations beyond the study's participants. Bias may arise from factors such as overrepresentation of certain demographic groups (e.g., age, gender, race), which can skew results and reduce the model's effectiveness across different populations from clinical settings.

\subsubsection{Methodological Challenges}
We classify the methodological challenges into the following categories: \textit{Unexplainability} of Deep learning methods, Longitudinal analysis, Computational complexity, and Transfer learning issues in medical settings.

\textbf{Unexplainability-Related Issues:} Overall, deep learning techniques are hardly explainable~\cite{yampolskiy2019unexplainability}, causing significant issues for applications that require transparency and interpretability, such as healthcare and cognitive impairment studies~\cite{murdoch2019definitions, molnar2020interpretable}. Primarily, certain modalities, such as visual ones (see Sec~\ref{subsec.Visual}), are still being explored, and without explainable indicators, the conclusions from these studies may be unclear. Robust validation of public datasets is essential for reliable detection of cognitive impairments. Likewise, despite the high detection performance reported by many research (see Sec.~\ref{sec.Performance}), it is crucial, from a medical standpoint, to identify and explain the indicators of cognitive conditions. 

\textbf{Longitudinal Analysis Issue:} Longitudinal studies track features over time, offering crucial insights into the progression of cognitive impairment from a neurological perspective. However, datasets, such as the ADReSS, capture only a single recording per subject~\cite{luz_alzheimers_2020}. In contrast, datasets like the Pitt Corpus offer a more comprehensive view over several years~\cite{becker1994natural}. Besides, most studies proposed to use a non-invasive modality for detecting AD (see Sec.~\ref{sec.Modeling and Modalities}), do not consider the timing of when data was collected. 

\textbf{Computational Complexity:} Many studies on cognitive detection systems using non-invasive data prioritize cognitive status detection over computational complexity. These studies harness various DL methods in their detection systems regardless of the computational complexities. For instance, Syed~\etal~\cite{syed_automated_2020} enhanced the detection performance of their approach by fusing five LLMs. However, this performance achievement consequently came at the cost of computational resources. Meanwhile, XnODR/XnIDR takes XNOR operation to lower the computational complexity, but they cause information loss for high-resolution images~\cite{sun2023xnodr}. Therefore, DL studies generally consider the computational complexities a key factor for comparison.

\textbf{Transfer Learning-Related Issues:} Transfer learning in deep learning approaches has shown promise in handling complex tasks~\cite{khan2023secure, zhu2023transfer, kheddar2023deep}. However, adapting these models to specific cognitive impairment datasets remains a challenge that requires further exploration due to some existing issues including data heterogeneity, and domain specificity. 

Data heterogeneity is an integral part of many cognitive impairment datasets, indicating these datasets provide more than one data modality (see Sec.\ref{sec.Datasets} for more details about existing datasets and the available data modalities). Usually, this heterogeneity makes it hard to apply previously trained models to these datasets~\cite{you2022ranking}. 

Likewise, in general, medical data (including cognitive data) might contain complex and nuances patterns that require more complex deep learning models to be captured. Hence, general-purpose models (\textit{e.g.} models trained on ImageNet~\cite{deng2009imagenet}) often show poor performance when applied to non-invasive cognitive data~\cite{wen2021rethinking}. In addition, proposing DL models that perform well across different datasets in a specific task requires a meticulous approach to model selection, architecture tuning, and robust evaluation strategies.

\subsubsection{Medical Adaptation Challenges}
Integration of deep learning detection systems in clinical settings faces a few technical challenges above the regulation of introducing a new detecting method. While there are free programs for preprocessing brain images~\cite{samann2022freesurfer}, non-invasive indicators are not yet well-represented in clinical settings. This gap poses a challenge for medical staff in utilizing these screening methods alongside existing diagnostic tools.

Although DL-detecting systems have achieved high detecting performance across various modalities, they still lack interpretability. Specifically, a reasonable understanding of the specific features that cause the model to reach such a detection level is absent. Consequently, Explainable AI (XAI) can bridge this gap. Incorporating XAI facilitates the adaptation of these DL methods as screening tools. As a result, medical personnel might have a better understanding and trust in the models~\cite{holzinger2017we, albahri2023systematic}.

\subsection{Future Research Directions}
We acknowledge that deep learning models are demonstrating competitive performance in detecting cognitive impairment. This article explores and investigates deep learning approaches, emphasizing their overall detecting performance over other techniques such as traditional machine learning algorithms. However, several future research directions can further enhance this field.

\textbf{Investigating Language-Agnostic Methods in Speech Analysis:}
Speech recordings and their transcripts are intrinsically coupled to the spoken language. Research on detection systems that integrate linguistic or acoustic features has predominantly focused on language-specific methods, often overlooking language-agnostic approaches. For instance, while gap filler words have been studied as linguistic features~\cite{yuan_pauses_2021}, more research is needed to understand their implications across different spoken languages comprehensively. Ultimately, adopting this method would also be beneficial in addressing the diversity challenge. 

\textbf{Proposing Qualitative Datasets:}
Although several publicly accessible datasets exist for research on cognitive impairment detection using non-invasive data, more comprehensive datasets should be created. The data collection should ensure a reliable and comprehensive representation of the population the impairment, and various data modalities by using consistent, efficient, and accurate methods and technologies. Consequently, enhancing the quality and diversity of datasets will support more robust model development.

Most current public datasets capture useful measurements of the studied participants. Nevertheless, from a data science perspective, some datasets lack common practices for creating datasets (i.e., data quality, sample size, class distribution, and baseline model), which could affect the adoption by deep learning researchers. For instance, Pitt Corpus and I-CONECT datasets have proposed comprehensive measurements with analyses of the participants; however, they still lack the introduction of baseline models and the evaluation subsets of the datasets. Implementing these practices ensures a fair comparison of any subsequent proposed deep learning models. The ADReSS dataset adopts these practices, which makes results on this dataset more reliable.

\textbf{Multi-Modal Cognitive Impairment Diagnosis:}
The ideal goal is a comprehensive analysis of all aspects of participants' indicators. This involves the development of AI systems that complement rather than replace physicians. These systems can serve as cost-effective, highly accurate screening tools, supporting medical professionals in their practice. Current diagnostic confidence is low and often relies on unverified data~\cite{better2024alzheimer}. Integrating advanced medical imaging such as MRI or CT with non-invasive indicators can improve diagnostic accuracy and explainability. 

Researchers can propose new multi-modal datasets and capture various modalities of the same subjects, including advanced brain imaging, speech, facial, and motoric mobility. These datasets will ultimately enhance the overall screening process and associate abnormal features together. Thus, it will help feature fusion. 

\textbf{Ethical and Explainable Considerations:}
As deep learning methods continue to evolve in healthcare, addressing ethical and privacy concerns remains crucial. Achieving a balance between maximizing benefits and minimizing risks is essential for the responsible use of AI in cognitive impairment detection. While deep learning detection methods necessitate training and validation of patient data, certain approaches help keep the data decentralized during these processes. \textit{Federated Learning}, for example, is one of the foremost algorithms in training deep learning models without centralizing the data~\cite{zhang2021survey}. By preserving patients' privacy, deep learning models can still learn and explain the desired features during training.

By addressing these challenges and exploring future research directions, we can harness the full potential of non-invasive modalities and deep learning models to improve cognitive impairment detection and ultimately enhance the quality of life for the aging population.

\section{Conclusion} \label{sec.Conclusion}

In this paper, we explored using deep learning methods as a detection method of cognitive impairment by using non-invasive data collection. Given that cognitive decline is on a large scale due to the aging population, it is crucial to adopt cost-effective screening tools to start a treatment plan as early as possible. We discussed the non-invasive indicators of cognitive decline and their support from the medical perspective. Specifically, these indicators were categorized into \textit{Speech}, \textit{Facial}, and \textit{Motoric Mobility} indicators. In addition, we highlighted the significant progress made in leveraging deep learning techniques to analyze various non-invasive data sources for cognitive impairment detection. 

Notably, speech-based methods, including acoustic and linguistic analyses, have shown leading outcomes in identifying subtle changes associated with cognitive decline. Particularly, the utilization of SOTA models from computer vision and natural language processing in extracting features and classifying conditions allows for a more nuanced understanding of complex datasets. Meanwhile, other modalities have demonstrated the potential to provide valuable insights into cognitive status detection. 

Although reviewed studies showed promising potential for adopting deep learning methods, several challenges hinder further systems improvements. We discussed a range of obstacles in detecting cognitive status, focusing on challenges from three main perspectives: data, methodological, and clinical adaptation. Standardization, diversity, and accessibility are among the data-related challenges whereas unexplainability, temporal analysis, and computation complexity are related to methodological adaptation of the studies. We then suggested methods and solutions to overcome these challenges. Ultimately, we proposed future research directions that can help the development of such systems.

\bibliographystyle{unsrt}

\bibliography{refs}

\begin{thebibliography}{100}

\bibitem{better2024alzheimer}
MAPPING~A BETTER.
\newblock alzheimer’s disease facts and figures.
\newblock {\em Alzheimer’s Dement}, 20:3708--3821, 2024.

\bibitem{pressman2024alzheimer}
Peter~S Pressman.
\newblock Alzheimer's disease.
\newblock 2024.

\bibitem{shamshirband2021review}
Shahab Shamshirband, Mahdis Fathi, Abdollah Dehzangi, Anthony~Theodore Chronopoulos, and Hamid Alinejad-Rokny.
\newblock A review on deep learning approaches in healthcare systems: Taxonomies, challenges, and open issues.
\newblock {\em Journal of Biomedical Informatics}, 113:103627, 2021.

\bibitem{haug2023artificial}
Charlotte~J Haug and Jeffrey~M Drazen.
\newblock Artificial intelligence and machine learning in clinical medicine, 2023.
\newblock {\em New England Journal of Medicine}, 388(13):1201--1208, 2023.

\bibitem{nevler2024changes}
Naomi Nevler, Sunghye Cho, Katheryn~AQ Cousins, Sharon Ash, Christopher~A Olm, Sanjana Shellikeri, Galit Agmon, Carmen Gonzalez-Recober, Sharon~X Xie, Megan~S Barker, et~al.
\newblock Changes in digital speech measures in asymptomatic carriers of pathogenic variants associated with frontotemporal degeneration.
\newblock {\em Neurology}, 102(2):e207926, 2024.

\bibitem{petti2023generalizability}
Ulla Petti, Simon Baker, Anna Korhonen, and Jessica Robin.
\newblock The generalizability of longitudinal changes in speech before alzheimer’s disease diagnosis.
\newblock {\em Journal of Alzheimer's Disease}, 92(2):547--564, 2023.

\bibitem{khan2021machine}
Protima Khan, Md~Fazlul Kader, SM~Riazul Islam, Aisha~B Rahman, Md~Shahriar Kamal, Masbah~Uddin Toha, and Kyung-Sup Kwak.
\newblock Machine learning and deep learning approaches for brain disease diagnosis: principles and recent advances.
\newblock {\em Ieee Access}, 9:37622--37655, 2021.

\bibitem{khojaste2022deep}
M~Khojaste-Sarakhsi, Seyedhamidreza~Shahabi Haghighi, SMT~Fatemi Ghomi, and Elena Marchiori.
\newblock Deep learning for alzheimer's disease diagnosis: A survey.
\newblock {\em Artificial intelligence in medicine}, 130:102332, 2022.

\bibitem{cummings2023lecanemab}
J~Cummings, L~Apostolova, GD~Rabinovici, A~Atri, P~Aisen, S~Greenberg, S~Hendrix, D~Selkoe, M~Weiner, RC~Petersen, et~al.
\newblock Lecanemab: appropriate use recommendations.
\newblock {\em The journal of prevention of Alzheimer's disease}, 10(3):362--377, 2023.

\bibitem{arciniegas2013behavioral}
David~B Arciniegas, C~Alan Anderson, and Christopher~M Filley.
\newblock {\em Behavioral neurology \& neuropsychiatry}.
\newblock Cambridge University Press, 2013.

\bibitem{becker1994natural}
James~T Becker, Fran{\c{c}}ois Boiler, Oscar~L Lopez, Judith Saxton, and Karen~L McGonigle.
\newblock The natural history of alzheimer's disease: description of study cohort and accuracy of diagnosis.
\newblock {\em Archives of neurology}, 51(6):585--594, 1994.

\bibitem{mugler2018differential}
Emily~M Mugler, Matthew~C Tate, Karen Livescu, Jessica~W Templer, Matthew~A Goldrick, and Marc~W Slutzky.
\newblock Differential representation of articulatory gestures and phonemes in precentral and inferior frontal gyri.
\newblock {\em Journal of Neuroscience}, 38(46):9803--9813, 2018.

\bibitem{mahon2022voice}
Elizabeth Mahon and Margie~E Lachman.
\newblock Voice biomarkers as indicators of cognitive changes in middle and later adulthood.
\newblock {\em Neurobiology of aging}, 119:22--35, 2022.

\bibitem{galluzzi2018aging}
Francesca Galluzzi and Werner Garavello.
\newblock The aging voice: a systematic review of presbyphonia.
\newblock {\em European Geriatric Medicine}, 9:559--570, 2018.

\bibitem{vaca2015aging}
Miguel Vaca, Elena Mora, and Ignacio Cobeta.
\newblock The aging voice: influence of respiratory and laryngeal changes.
\newblock {\em Otolaryngology--Head and Neck Surgery}, 153(3):409--413, 2015.

\bibitem{baevski2020wav2vec}
Alexei Baevski, Yuhao Zhou, Abdelrahman Mohamed, and Michael Auli.
\newblock wav2vec 2.0: A framework for self-supervised learning of speech representations.
\newblock {\em Advances in neural information processing systems}, 33:12449--12460, 2020.

\bibitem{hershey2017cnn}
Shawn Hershey, Sourish Chaudhuri, Daniel~PW Ellis, Jort~F Gemmeke, Aren Jansen, R~Channing Moore, Manoj Plakal, Devin Platt, Rif~A Saurous, Bryan Seybold, et~al.
\newblock Cnn architectures for large-scale audio classification.
\newblock In {\em 2017 ieee international conference on acoustics, speech and signal processing (icassp)}, pages 131--135. IEEE, 2017.

\bibitem{verfaillie2018more}
Sander~CJ Verfaillie, Rosalinde~ER Slot, Ellen Dicks, Niels~D Prins, Jozefien~M Overbeek, Charlotte~E Teunissen, Philip Scheltens, Frederik Barkhof, Wiesje~M van~der Flier, and Betty~M Tijms.
\newblock A more randomly organized grey matter network is associated with deteriorating language and global cognition in individuals with subjective cognitive decline.
\newblock {\em Human brain mapping}, 39(8):3143--3151, 2018.

\bibitem{mesulam2001primary}
M-Marsel Mesulam.
\newblock Primary progressive aphasia.
\newblock {\em Annals of neurology}, 49(4):425--432, 2001.

\bibitem{petersen2016mild}
Ronald~C Petersen.
\newblock Mild cognitive impairment.
\newblock {\em CONTINUUM: lifelong Learning in Neurology}, 22(2):404--418, 2016.

\bibitem{pressman2023incongruences}
Peter~S Pressman, Kuan~Hua Chen, James Casey, Stefan Sillau, Heidi~J Chial, Christopher~M Filley, Bruce~L Miller, and Robert~W Levenson.
\newblock Incongruences between facial expression and self-reported emotional reactivity in frontotemporal dementia and related disorders.
\newblock {\em The Journal of neuropsychiatry and clinical neurosciences}, 35(2):192--201, 2023.

\bibitem{pressman2014diagnosis}
Peter~S Pressman and Bruce~L Miller.
\newblock Diagnosis and management of behavioral variant frontotemporal dementia.
\newblock {\em Biological psychiatry}, 75(7):574--581, 2014.

\bibitem{pressman2017relative}
Peter Pressman, Kelly Gola, Suzanne Shdo, Bruce Miller, and Katherine Rankin.
\newblock Relative preservation of affect recognition in posterior cortical atrophy (s18. 006).
\newblock {\em Neurology}, 88(16\_supplement):S18--006, 2017.

\bibitem{fei_novel_2022}
Zixiang Fei, Erfu Yang, Leijian Yu, Xia Li, Huiyu Zhou, and Wenju Zhou.
\newblock A {Novel} deep neural network-based emotion analysis system for automatic detection of mild cognitive impairment in the elderly.
\newblock {\em Neurocomputing}, 468:306--316, January 2022.

\bibitem{katz2016early}
Maya Katz, Peter Pressman, and Bradley~F Boeve.
\newblock Early clinical features of the parkinsonian-related dementias.
\newblock {\em The Behavioral Neurology of Dementia}, pages 232--244, 2016.

\bibitem{harada2013normal}
Caroline~N Harada, Marissa C~Natelson Love, and Kristen~L Triebel.
\newblock Normal cognitive aging.
\newblock {\em Clinics in geriatric medicine}, 29(4):737--752, 2013.

\bibitem{rentz2021association}
Dorene~M Rentz, Kathryn~V Papp, Danielle~V Mayblyum, Justin~S Sanchez, Hannah Klein, William Souillard-Mandar, Reisa~A Sperling, and Keith~A Johnson.
\newblock Association of digital clock drawing with pet amyloid and tau pathology in normal older adults.
\newblock {\em Neurology}, 96(14):e1844--e1854, 2021.

\bibitem{nasreddine2005montreal}
Ziad~S Nasreddine, Natalie~A Phillips, Val{\'e}rie B{\'e}dirian, Simon Charbonneau, Victor Whitehead, Isabelle Collin, Jeffrey~L Cummings, and Howard Chertkow.
\newblock The montreal cognitive assessment, moca: a brief screening tool for mild cognitive impairment.
\newblock {\em Journal of the American Geriatrics Society}, 53(4):695--699, 2005.

\bibitem{cilia_handwriting-based_2021}
Nicole~Dalia Cilia, Claudio De~Stefano, Francesco Fontanella, and Alessandra Scotto~di Freca.
\newblock Handwriting-{Based} {Classifier} {Combination} for {Cognitive} {Impairment} {Prediction}.
\newblock In Alberto Del~Bimbo, Rita Cucchiara, Stan Sclaroff, Giovanni~Maria Farinella, Tao Mei, Marco Bertini, Hugo~Jair Escalante, and Roberto Vezzani, editors, {\em Pattern {Recognition}. {ICPR} {International} {Workshops} and {Challenges}}, Lecture {Notes} in {Computer} {Science}, pages 587--599, Cham, 2021. Springer International Publishing.

\bibitem{poor2024multimodal}
Farida~Far Poor, Hiroko~H Dodge, and Mohammad~H Mahoor.
\newblock A multimodal cross-transformer-based model to predict mild cognitive impairment using speech, language and vision.
\newblock {\em Computers in Biology and Medicine}, 182:109199, 2024.

\bibitem{herd2014cohort}
Pamela Herd, Deborah Carr, and Carol Roan.
\newblock Cohort profile: Wisconsin longitudinal study (wls).
\newblock {\em International journal of epidemiology}, 43(1):34--41, 2014.

\bibitem{luz_alzheimers_2020}
Saturnino Luz, Fasih Haider, Sofia de~la Fuente, Davida Fromm, and Brian MacWhinney.
\newblock Alzheimer's {Dementia} {Recognition} through {Spontaneous} {Speech}: {The} {ADReSS} {Challenge}, August 2020.
\newblock arXiv:2004.06833 [cs, eess, stat].

\bibitem{luz_detecting_2021}
Saturnino Luz, Fasih Haider, Sofia de~la Fuente, Davida Fromm, and Brian MacWhinney.
\newblock Detecting cognitive decline using speech only: {The} {ADReSSo} {Challenge}, March 2021.
\newblock arXiv:2104.09356 [cs, eess].

\bibitem{ying_multimodal_2023}
Yangwei Ying, Tao Yang, and Hong Zhou.
\newblock Multimodal fusion for alzheimer’s disease recognition.
\newblock {\em Applied Intelligence}, 53(12):16029--16040, June 2023.

\bibitem{nishikawa_detecting_2021}
Kazu Nishikawa, Rin Hirakawa, Hideaki Kawano, Kenichi Nakashi, and Yoshihisa Nakatoh.
\newblock Detecting {System} {Alzheimer}’s {Dementia} by 1d {CNN}-{LSTM} in {Japanese} {Speech}.
\newblock In {\em 2021 {IEEE} {International} {Conference} on {Consumer} {Electronics} ({ICCE})}, pages 1--3, January 2021.
\newblock ISSN: 2158-4001.

\bibitem{you_alzheimers_2020}
Zeng You, Runhao Zeng, Xiaoyong Lan, Huixia Ren, Zhiyang You, Xue Shi, Shipeng Zhao, Yi~Guo, Xin Jiang, and Xiping Hu.
\newblock Alzheimer's {Disease} {Classification} {With} a {Cascade} {Neural} {Network}.
\newblock {\em Frontiers in Public Health}, 8, 2020.

\bibitem{you_alzheimers_2021}
Zhiyang You, Zeng You, Yilong Li, Shipeng Zhao, Huixia Ren, and Xiping Hu.
\newblock Alzheimer's {Disease} {Distinction} {Based} {On} {Gait} {Feature} {Analysis}.
\newblock In {\em 2020 {IEEE} {International} {Conference} on {E}-health {Networking}, {Application} \& {Services} ({HEALTHCOM})}, pages 1--6, March 2021.

\bibitem{zuo_deep_2023}
Fangyu Zuo, Peiguang Jing, Jinglin Sun, Jizhong, Duan, Yong Ji, and Yu~Liu.
\newblock Deep {Learning}-based {Eye}-{Tracking} {Analysis} for {Diagnosis} of {Alzheimer}'s {Disease} {Using} {3D} {Comprehensive} {Visual} {Stimuli}, March 2023.
\newblock arXiv:2303.06868 [cs, eess].

\bibitem{khan_stacked_2022}
Yusera~Farooq Khan, Baijnath Kaushik, Mohammad Khalid~Imam Rahmani, and Md~Ezaz Ahmed.
\newblock Stacked {Deep} {Dense} {Neural} {Network} {Model} to {Predict} {Alzheimer}’s {Dementia} {Using} {Audio} {Transcript} {Data}.
\newblock {\em IEEE Access}, 10:32750--32765, 2022.
\newblock Conference Name: IEEE Access.

\bibitem{orimaye_deep_2018}
Sylvester~Olubolu Orimaye, Jojo Sze-Meng Wong, and Chee~Piau Wong.
\newblock Deep language space neural network for classifying mild cognitive impairment and {Alzheimer}-type dementia.
\newblock {\em PLOS ONE}, 13(11):e0205636, November 2018.
\newblock Publisher: Public Library of Science.

\bibitem{fritsch_automatic_2019}
Julian Fritsch, Sebastian Wankerl, and Elmar Nöth.
\newblock Automatic {Diagnosis} of {Alzheimer}’s {Disease} {Using} {Neural} {Network} {Language} {Models}.
\newblock In {\em {ICASSP} 2019 - 2019 {IEEE} {International} {Conference} on {Acoustics}, {Speech} and {Signal} {Processing} ({ICASSP})}, pages 5841--5845, May 2019.
\newblock ISSN: 2379-190X.

\bibitem{guo_crossing_2021}
Yue Guo, Changye Li, Carol Roan, Serguei Pakhomov, and Trevor Cohen.
\newblock Crossing the “{Cookie} {Theft}” {Corpus} {Chasm}: {Applying} {What} {BERT} {Learns} {From} {Outside} {Data} to the {ADReSS} {Challenge} {Dementia} {Detection} {Task}.
\newblock {\em Frontiers in Computer Science}, 3, 2021.

\bibitem{cummins_comparison_2020}
Nicholas Cummins, Yilin Pan, Zhao Ren, Julian Fritsch, Venkata~Srikanth Nallanthighal, Heidi Christensen, Daniel Blackburn, Björn~W. Schuller, Mathew Magimai-Doss, Helmer Strik, and Aki Härmä.
\newblock A {Comparison} of {Acoustic} and {Linguistics} {Methodologies} for {Alzheimer}’s {Dementia} {Recognition}.
\newblock In {\em Interspeech 2020}, pages 2182--2186. ISCA, October 2020.

\bibitem{koo_exploiting_2020}
Junghyun Koo, Jie~Hwan Lee, Jaewoo Pyo, Yujin Jo, and Kyogu Lee.
\newblock Exploiting {Multi}-{Modal} {Features} from {Pre}-{Trained} {Networks} for {Alzheimer}’s {Dementia} {Recognition}.
\newblock In {\em Interspeech 2020}, pages 2217--2221. ISCA, October 2020.

\bibitem{syed_automated_2021}
Zafi~Sherhan Syed, Muhammad Shehram~Shah Syed, Margaret Lech, and Elena Pirogova.
\newblock Automated {Recognition} of {Alzheimer}’s {Dementia} {Using} {Bag}-of-{Deep}-{Features} and {Model} {Ensembling}.
\newblock {\em IEEE Access}, 9:88377--88390, 2021.
\newblock Conference Name: IEEE Access.

\bibitem{pan_using_2021}
Yilin Pan, Bahman Mirheidari, Jennifer~M. Harris, Jennifer~C. Thompson, Matthew Jones, Julie~S. Snowden, Daniel Blackburn, and Heidi Christensen.
\newblock Using the {Outputs} of {Different} {Automatic} {Speech} {Recognition} {Paradigms} for {Acoustic}- and {BERT}-{Based} {Alzheimer}\&\#8217;s {Dementia} {Detection} {Through} {Spontaneous} {Speech}.
\newblock In {\em Interspeech 2021}, pages 3810--3814. ISCA, August 2021.

\bibitem{gauder_alzheimer_2021}
Lara Gauder, Leonardo Pepino, Luciana Ferrer, and Pablo Riera.
\newblock Alzheimer {Disease} {Recognition} {Using} {Speech}-{Based} {Embeddings} {From} {Pre}-{Trained} {Models}.
\newblock In {\em Interspeech 2021}, pages 3795--3799. ISCA, August 2021.

\bibitem{wang_modular_2021}
Ning Wang, Yupeng Cao, Shuai Hao, Zongru Shao, and K.P. Subbalakshmi.
\newblock Modular {Multi}-{Modal} {Attention} {Network} for {Alzheimer}\&\#8217;s {Disease} {Detection} {Using} {Patient} {Audio} and {Language} {Data}.
\newblock In {\em Interspeech 2021}, pages 3835--3839. ISCA, August 2021.

\bibitem{kishimoto_project_2020}
Taishiro Kishimoto, Akihiro Takamiya, Kuo-ching Liang, Kei Funaki, Takanori Fujita, Momoko Kitazawa, Michitaka Yoshimura, Yuki Tazawa, Toshiro Horigome, Yoko Eguchi, Toshiaki Kikuchi, Masayuki Tomita, Shogyoku Bun, Junichi Murakami, Brian Sumali, Tifani Warnita, Aiko Kishi, Mizuki Yotsui, Hiroyoshi Toyoshiba, Yasue Mitsukura, Koichi Shinoda, Yasubumi Sakakibara, and Masaru Mimura.
\newblock The project for objective measures using computational psychiatry technology ({PROMPT}): {Rationale}, design, and methodology.
\newblock {\em Contemporary Clinical Trials Communications}, 19:100649, September 2020.

\bibitem{rodrigues_makiuchi_speech_2021}
Mariana Rodrigues~Makiuchi, Tifani Warnita, Nakamasa Inoue, Koichi Shinoda, Michitaka Yoshimura, Momoko Kitazawa, Kei Funaki, Yoko Eguchi, and Taishiro Kishimoto.
\newblock Speech {Paralinguistic} {Approach} for {Detecting} {Dementia} {Using} {Gated} {Convolutional} {Neural} {Network}.
\newblock {\em IEICE Transactions on Information and Systems}, E104.D(11):1930--1940, November 2021.

\bibitem{zheng_detecting_2023}
Chuheng Zheng, Mondher Bouazizi, Tomoaki Ohtsuki, Momoko Kitazawa, Toshiro Horigome, and Taishiro Kishimoto.
\newblock Detecting {Dementia} from {Face}-{Related} {Features} with {Automated} {Computational} {Methods}.
\newblock {\em Bioengineering}, 10(7):862, July 2023.

\bibitem{dodge2024internet}
Hiroko~H Dodge, Kexin Yu, Chao-Yi Wu, Patrick~J Pruitt, Meysam Asgari, Jeffrey~A Kaye, Benjamin~M Hampstead, Laura Struble, Kathleen Potempa, Peter Lichtenberg, et~al.
\newblock Internet-based conversational engagement randomized controlled clinical trial (i-conect) among socially isolated adults 75+ years old with normal cognition or mild cognitive impairment: Topline results.
\newblock {\em The Gerontologist}, 64(4):gnad147, 2024.

\bibitem{sun_mc-vivit_2024}
Jian Sun, Hiroko~Hayama Dodge, and Mohammad~H. Mahoor.
\newblock {MC}-{ViViT}: {Multi}-branch {Classifier}-{ViViT} to detect {Mild} {Cognitive} {Impairment} in older adults using facial videos.
\newblock {\em Expert Systems with Applications}, 238:121929, March 2024.

\bibitem{fard2024linguistic}
Ali~Pourramezan Fard, Mohammad~H Mahoor, Muath Alsuhaibani, and Hiroko~H Dodge.
\newblock Linguistic-based mild cognitive impairment detection using informative loss.
\newblock {\em Computers in Biology and Medicine}, page 108606, 2024.

\bibitem{alsuhaibani2024mild}
Muath Alsuhaibani, Hiroko~H Dodge, and Mohammad~H Mahoor.
\newblock Mild cognitive impairment detection from facial video interviews by applying spatial-to-temporal attention module.
\newblock {\em Expert Systems with Applications}, page 124185, 2024.

\bibitem{bringas_alzheimers_2020}
Santos Bringas, Sergio Salomón, Rafael Duque, Carmen Lage, and José~Luis Montaña.
\newblock Alzheimer’s {Disease} stage identification using deep learning models.
\newblock {\em Journal of Biomedical Informatics}, 109:103514, September 2020.

\bibitem{aoki_early_2019}
Kota Aoki, Trung~Thanh Ngo, Ikuhisa Mitsugami, Fumio Okura, Masataka Niwa, Yasushi Makihara, Yasushi Yagi, and Hiroaki Kazui.
\newblock Early {Detection} of {Lower} {MMSE} {Scores} in {Elderly} {Based} on {Dual}-{Task} {Gait}.
\newblock {\em IEEE Access}, 7:40085--40094, 2019.
\newblock Conference Name: IEEE Access.

\bibitem{shahzad_automated_2022}
Ahsan Shahzad, Aresh Dadlani, Hyeonil Lee, and Kiseon Kim.
\newblock Automated {Prescreening} of {Mild} {Cognitive} {Impairment} {Using} {Shank}-{Mounted} {Inertial} {Sensors} {Based} {Gait} {Biomarkers}.
\newblock {\em IEEE Access}, 10:15835--15844, 2022.
\newblock Conference Name: IEEE Access.

\bibitem{ghoraani_detection_2021}
Behnaz Ghoraani, Lillian~N. Boettcher, Murtadha~D. Hssayeni, Amie Rosenfeld, Magdalena~I. Tolea, and James~E. Galvin.
\newblock Detection of mild cognitive impairment and {Alzheimer}’s disease using dual-task gait assessments and machine learning.
\newblock {\em Biomedical Signal Processing and Control}, 64:102249, February 2021.

\bibitem{narasimhan_early_2020}
Rajaram Narasimhan, Muthukumaran G, Charles McGlade, and Anantha Ramakrishnan.
\newblock Early {Detection} of {Mild} {Cognitive} {Impairment} {Progression} {Using} {Non}-{Wearable} {Sensor} {Data} – a {Deep} {Learning} {Approach}.
\newblock In {\em 2020 {IEEE} {Bangalore} {Humanitarian} {Technology} {Conference} ({B}-{HTC})}, pages 1--6, October 2020.

\bibitem{el-yacoubi_aging_2019}
Mounîm~A. El-Yacoubi, Sonia Garcia-Salicetti, Christian Kahindo, Anne-Sophie Rigaud, and Victoria Cristancho-Lacroix.
\newblock From aging to early-stage {Alzheimer}'s: {Uncovering} handwriting multimodal behaviors by semi-supervised learning and sequential representation learning.
\newblock {\em Pattern Recognition}, 86:112--133, February 2019.

\bibitem{yu_internet-based_2021}
Kexin Yu, Katherine Wild, Kathleen Potempa, Benjamin~M. Hampstead, Peter~A. Lichtenberg, Laura~M. Struble, Patrick Pruitt, Elena~L. Alfaro, Jacob Lindsley, Mattie MacDonald, Jeffrey~A. Kaye, Lisa~C. Silbert, and Hiroko~H. Dodge.
\newblock The {Internet}-{Based} {Conversational} {Engagement} {Clinical} {Trial} ({I}-{CONECT}) in {Socially} {Isolated} {Adults} 75+ {Years} {Old}: {Randomized} {Controlled} {Trial} {Protocol} and {COVID}-19 {Related} {Study} {Modifications}.
\newblock {\em Frontiers in Digital Health}, 3, 2021.

\bibitem{he2016deep}
Kaiming He, Xiangyu Zhang, Shaoqing Ren, and Jian Sun.
\newblock Deep residual learning for image recognition.
\newblock In {\em Proceedings of the IEEE conference on computer vision and pattern recognition}, pages 770--778, 2016.

\bibitem{simonyan2014very}
Karen Simonyan and Andrew Zisserman.
\newblock Very deep convolutional networks for large-scale image recognition.
\newblock {\em arXiv preprint arXiv:1409.1556}, 2014.

\bibitem{sandler2018mobilenetv2}
Mark Sandler, Andrew Howard, Menglong Zhu, Andrey Zhmoginov, and Liang-Chieh Chen.
\newblock Mobilenetv2: Inverted residuals and linear bottlenecks.
\newblock In {\em Proceedings of the IEEE conference on computer vision and pattern recognition}, pages 4510--4520, 2018.

\bibitem{cho2014learning}
Kyunghyun Cho.
\newblock Learning phrase representations using rnn encoder-decoder for statistical machine translation.
\newblock {\em arXiv preprint arXiv:1406.1078}, 2014.

\bibitem{hochreiter1997long}
S~Hochreiter.
\newblock Long short-term memory.
\newblock {\em Neural Computation MIT-Press}, 1997.

\bibitem{vaswani2017attention}
A~Vaswani.
\newblock Attention is all you need.
\newblock {\em Advances in Neural Information Processing Systems}, 2017.

\bibitem{radford2019language}
Alec Radford, Jeffrey Wu, Rewon Child, David Luan, Dario Amodei, Ilya Sutskever, et~al.
\newblock Language models are unsupervised multitask learners.
\newblock {\em OpenAI blog}, 1(8):9, 2019.

\bibitem{devlin2018bert}
Jacob Devlin, Ming-Wei Chang, Kenton Lee, and Kristina Toutanova.
\newblock Bert: Pre-training of deep bidirectional transformers for language understanding.
\newblock {\em arXiv preprint arXiv:1810.04805}, 2018.

\bibitem{dosovitskiy2020image}
Alexey Dosovitskiy, Lucas Beyer, Alexander Kolesnikov, Dirk Weissenborn, Xiaohua Zhai, Thomas Unterthiner, Mostafa Dehghani, Matthias Minderer, Georg Heigold, Sylvain Gelly, et~al.
\newblock An image is worth 16x16 words: Transformers for image recognition at scale.
\newblock {\em arXiv preprint arXiv:2010.11929}, 2020.

\bibitem{kumar_dementia_2022}
M.~Rupesh Kumar, Susmitha Vekkot, S.~Lalitha, Deepa Gupta, Varasiddhi~Jayasuryaa Govindraj, Kamran Shaukat, Yousef~Ajami Alotaibi, and Mohammed Zakariah.
\newblock Dementia {Detection} from {Speech} {Using} {Machine} {Learning} and {Deep} {Learning} {Architectures}.
\newblock {\em Sensors}, 22(23):9311, January 2022.
\newblock Number: 23 Publisher: Multidisciplinary Digital Publishing Institute.

\bibitem{liu_detecting_2021}
Zhaoci Liu, Zhiqiang Guo, Zhenhua Ling, and Yunxia Li.
\newblock Detecting {Alzheimer}’s {Disease} from {Speech} {Using} {Neural} {Networks} with {Bottleneck} {Features} and {Data} {Augmentation}.
\newblock In {\em {ICASSP} 2021 - 2021 {IEEE} {International} {Conference} on {Acoustics}, {Speech} and {Signal} {Processing} ({ICASSP})}, pages 7323--7327, June 2021.
\newblock ISSN: 2379-190X.

\bibitem{meghanani_exploration_2021}
Amit Meghanani, Anoop C.~S., and A.~G. Ramakrishnan.
\newblock An {Exploration} of {Log}-{Mel} {Spectrogram} and {MFCC} {Features} for {Alzheimer}’s {Dementia} {Recognition} from {Spontaneous} {Speech}.
\newblock In {\em 2021 {IEEE} {Spoken} {Language} {Technology} {Workshop} ({SLT})}, pages 670--677, January 2021.

\bibitem{warnita_detecting_2018}
Tifani Warnita, Nakamasa Inoue, and Koichi Shinoda.
\newblock Detecting {Alzheimer}’s {Disease} {Using} {Gated} {Convolutional} {Neural} {Network} from {Audio} {Data}.
\newblock In {\em Interspeech 2018}, pages 1706--1710. ISCA, September 2018.

\bibitem{syed_static_2020}
Muhammad Shehram~Shah Syed, Zafi Sherhan, Elena Pirogova, and Margaret Lech.
\newblock Static vs. {Dynamic} {Modelling} of {Acoustic} {Speech} {Features} for {Detection} of {Dementia}.
\newblock {\em International Journal of Advanced Computer Science and Applications}, 11(10), 2020.

\bibitem{chlasta_towards_2021}
Karol Chlasta and Krzysztof Wołk.
\newblock Towards {Computer}-{Based} {Automated} {Screening} of {Dementia} {Through} {Spontaneous} {Speech}.
\newblock {\em Frontiers in Psychology}, 11, 2021.

\bibitem{nishikawa_machine_2022}
Kazu Nishikawa, Kuwahara Akihiro, Rin Hirakawa, Hideaki Kawano, and Yoshihisa Nakatoh.
\newblock Machine learning model for discrimination of mild dementia patients using acoustic features.
\newblock {\em Cognitive Robotics}, 2:21--29, January 2022.

\bibitem{pranav_early_2023}
G.~Pranav, K.~Varsha, and K.~S. Gayathri.
\newblock Early {Alzheimer} {Detection} {Through} {Speech} {Analysis} and {Vision} {Transformer} {Approach}.
\newblock In Anand~Kumar M, Bharathi~Raja Chakravarthi, Bharathi B, Colm O’Riordan, Hema Murthy, Thenmozhi Durairaj, and Thomas Mandl, editors, {\em Speech and {Language} {Technologies} for {Low}-{Resource} {Languages}}, Communications in {Computer} and {Information} {Science}, pages 265--276, Cham, 2023. Springer International Publishing.

\bibitem{bertini_automatic_2022}
Flavio Bertini, Davide Allevi, Gianluca Lutero, Laura Calzà, and Danilo Montesi.
\newblock An automatic {Alzheimer}’s disease classifier based on spontaneous spoken {English}.
\newblock {\em Computer Speech \& Language}, 72:101298, March 2022.

\bibitem{bertini_automatic_2021}
Flavio Bertini, Davide Allevi, Gianluca Lutero, Danilo Montesi, and Laura Calzà.
\newblock Automatic {Speech} {Classifier} for {Mild} {Cognitive} {Impairment} and {Early} {Dementia}.
\newblock {\em ACM Transactions on Computing for Healthcare}, 3(1):8:1--8:11, October 2021.

\bibitem{pan_acoustic_2020}
Yilin Pan, Bahman Mirheidari, Zehai Tu, Ronan O’Malley, Traci Walker, Annalena Venneri, Markus Reuber, Daniel Blackburn, and Heidi Christensen.
\newblock Acoustic {Feature} {Extraction} with {Interpretable} {Deep} {Neural} {Network} for {Neurodegenerative} {Related} {Disorder} {Classification}.
\newblock In {\em Interspeech 2020}, pages 4806--4810. ISCA, October 2020.

\bibitem{xue_detection_2021}
Chonghua Xue, Cody Karjadi, Ioannis~Ch. Paschalidis, Rhoda Au, and Vijaya~B. Kolachalama.
\newblock Detection of dementia on voice recordings using deep learning: a {Framingham} {Heart} {Study}.
\newblock {\em Alzheimer's Research \& Therapy}, 13(1):146, August 2021.

\bibitem{zolnoori_adscreen_2023}
Maryam Zolnoori, Ali Zolnour, and Maxim Topaz.
\newblock {ADscreen}: {A} speech processing-based screening system for automatic identification of patients with {Alzheimer}'s disease and related dementia.
\newblock {\em Artificial Intelligence in Medicine}, 143:102624, September 2023.

\bibitem{lin_automatic_2022}
Sheng-Ya Lin, Ho-Ling Chang, Jwu-Jia Hwang, Thiri Wai, Yu-Ling Chang, and Li-Chen Fu.
\newblock Automatic {Audio}-based {Screening} {System} for {Alzheimer}’s {Disease} {Detection}.
\newblock In {\em 2022 {IEEE} {International} {Conference} on {Systems}, {Man}, and {Cybernetics} ({SMC})}, pages 2770--2775, October 2022.
\newblock ISSN: 2577-1655.

\bibitem{syed_automated_2020}
Muhammad Shehram~Shah Syed, Zafi~Sherhan Syed, Margaret Lech, and Elena Pirogova.
\newblock Automated {Screening} for {Alzheimer}’s {Dementia} {Through} {Spontaneous} {Speech}.
\newblock In {\em Interspeech 2020}, pages 2222--2226. ISCA, October 2020.

\bibitem{eyben2015geneva}
Florian Eyben, Klaus~R Scherer, Bj{\"o}rn~W Schuller, Johan Sundberg, Elisabeth Andr{\'e}, Carlos Busso, Laurence~Y Devillers, Julien Epps, Petri Laukka, Shrikanth~S Narayanan, et~al.
\newblock The geneva minimalistic acoustic parameter set (gemaps) for voice research and affective computing.
\newblock {\em IEEE transactions on affective computing}, 7(2):190--202, 2015.

\bibitem{pappagari_automatic_2021}
Raghavendra Pappagari, Jaejin Cho, Sonal Joshi, Laureano Moro-Velázquez, Piotr Żelasko, Jesús Villalba, and Najim Dehak.
\newblock Automatic {Detection} and {Assessment} of {Alzheimer} {Disease} {Using} {Speech} and {Language} {Technologies} in {Low}-{Resource} {Scenarios}.
\newblock In {\em Interspeech 2021}, pages 3825--3829. ISCA, August 2021.

\bibitem{campbell_alzheimers_2020}
Edward~L. Campbell, Laura Docío-Fernández, Javier~Jiménez Raboso, and Carmen García-Mateo.
\newblock Alzheimer's {Dementia} {Detection} from {Audio} and {Text} {Modalities}, August 2020.
\newblock arXiv:2008.04617 [cs, eess].

\bibitem{eyben2010opensmile}
Florian Eyben, Martin W{\"o}llmer, and Bj{\"o}rn Schuller.
\newblock Opensmile: the munich versatile and fast open-source audio feature extractor.
\newblock In {\em Proceedings of the 18th ACM international conference on Multimedia}, pages 1459--1462, 2010.

\bibitem{mcfee2015librosa}
Brian McFee, Colin Raffel, Dawen Liang, Daniel~PW Ellis, Matt McVicar, Eric Battenberg, and Oriol Nieto.
\newblock librosa: Audio and music signal analysis in python.
\newblock In {\em SciPy}, pages 18--24, 2015.

\bibitem{mehrish2023review}
Ambuj Mehrish, Navonil Majumder, Rishabh Bharadwaj, Rada Mihalcea, and Soujanya Poria.
\newblock A review of deep learning techniques for speech processing.
\newblock {\em Information Fusion}, 99:101869, 2023.

\bibitem{pancoast2012bag}
Stephanie Pancoast and Murat Akbacak.
\newblock Bag-of-audio-words approach for multimedia event classification.
\newblock In {\em Interspeech}, pages 2105--2108, 2012.

\bibitem{snyder2018x}
David Snyder, Daniel Garcia-Romero, Gregory Sell, Daniel Povey, and Sanjeev Khudanpur.
\newblock X-vectors: Robust dnn embeddings for speaker recognition.
\newblock In {\em 2018 IEEE international conference on acoustics, speech and signal processing (ICASSP)}, pages 5329--5333. IEEE, 2018.

\bibitem{li_leveraging_2023}
Jinchao Li, Kaitao Song, Junan Li, Bo~Zheng, Dongsheng Li, Xixin Wu, Xunying Liu, and Helen Meng.
\newblock Leveraging {Pretrained} {Representations} {With} {Task}-{Related} {Keywords} for {Alzheimer}’s {Disease} {Detection}.
\newblock In {\em {ICASSP} 2023 - 2023 {IEEE} {International} {Conference} on {Acoustics}, {Speech} and {Signal} {Processing} ({ICASSP})}, pages 1--5, June 2023.

\bibitem{searle_comparing_2020}
Thomas Searle, Zina Ibrahim, and Richard Dobson.
\newblock Comparing {Natural} {Language} {Processing} {Techniques} for {Alzheimer}’s {Dementia} {Prediction} in {Spontaneous} {Speech}.
\newblock In {\em Interspeech 2020}, pages 2192--2196. ISCA, October 2020.

\bibitem{meghanani_recognition_2021}
Amit Meghanani, C.~S. Anoop, and Angarai~Ganesan Ramakrishnan.
\newblock Recognition of {Alzheimer}’s {Dementia} {From} the {Transcriptions} of {Spontaneous} {Speech} {Using} {fastText} and {CNN} {Models}.
\newblock {\em Frontiers in Computer Science}, 3, 2021.

\bibitem{yuan_pauses_2021}
Jiahong Yuan, Xingyu Cai, Yuchen Bian, Zheng Ye, and Kenneth Church.
\newblock Pauses for {Detection} of {Alzheimer}’s {Disease}.
\newblock {\em Frontiers in Computer Science}, 2, 2021.

\bibitem{casanova_evaluating_2020}
Edresson Casanova, Marcos Treviso, Lilian Hübner, and Sandra Aluísio.
\newblock Evaluating {Sentence} {Segmentation} in {Different} {Datasets} of {Neuropsychological} {Language} {Tests} in {Brazilian} {Portuguese}.
\newblock In {\em Proceedings of the {Twelfth} {Language} {Resources} and {Evaluation} {Conference}}, pages 2605--2614, Marseille, France, May 2020. European Language Resources Association.

\bibitem{liu_improving_2022}
Ning Liu, Zhenming Yuan, and Qingfeng Tang.
\newblock Improving {Alzheimer}'s {Disease} {Detection} for {Speech} {Based} on {Feature} {Purification} {Network}.
\newblock {\em Frontiers in Public Health}, 9, 2022.

\bibitem{tsai_efficient_2021}
Austin Cheng-Yun Tsai, Sheng-Yi Hong, Li-Hung Yao, Wei-Der Chang, Li-Chen Fu, and Yu-Ling Chang.
\newblock An efficient context-aware screening system for {Alzheimer}'s disease based on neuropsychology test.
\newblock {\em Scientific Reports}, 11(1):18570, September 2021.
\newblock Number: 1 Publisher: Nature Publishing Group.

\bibitem{chen_attention-based_2019}
Jun Chen, Ji~Zhu, and Jieping Ye.
\newblock An {Attention}-{Based} {Hybrid} {Network} for {Automatic} {Detection} of {Alzheimer}’s {Disease} from {Narrative} {Speech}.
\newblock In {\em Interspeech 2019}, pages 4085--4089. ISCA, September 2019.

\bibitem{ai-atroshi_automated_2022}
Chiai AI-Atroshi, J.~Rene~Beulah, Kranthi~Kumar Singamaneni, C.~Pretty Diana~Cyril, S.~Neelakandan, and S.~Velmurugan.
\newblock Automated speech based evaluation of mild cognitive impairment and {Alzheimer}’s disease detection using with deep belief network model.
\newblock {\em International Journal of Healthcare Management}, 0(0):1--11, July 2022.
\newblock Publisher: Taylor \& Francis \_eprint: https://doi.org/10.1080/20479700.2022.2097764.

\bibitem{wen_revealing_2023}
Bingyang Wen, Ning Wang, Koduvayur Subbalakshmi, and Rajarathnam Chandramouli.
\newblock Revealing the {Roles} of {Part}-of-{Speech} {Taggers} in {Alzheimer} {Disease} {Detection}: {Scientific} {Discovery} {Using} {One}-{Intervention} {Causal} {Explanation}.
\newblock {\em JMIR Formative Research}, 7(1):e36590, May 2023.
\newblock Company: JMIR Formative Research Distributor: JMIR Formative Research Institution: JMIR Formative Research Label: JMIR Formative Research Publisher: JMIR Publications Inc., Toronto, Canada.

\bibitem{alkenani_predicting_2021}
Ahmed~H. Alkenani, Yuefeng Li, Yue Xu, and Qing Zhang.
\newblock Predicting {Alzheimer}’s {Disease} from {Spoken} and {Written} {Language} {Using} {Fusion}-{Based} {Stacked} {Generalization}.
\newblock {\em Journal of Biomedical Informatics}, 118:103803, June 2021.

\bibitem{wang_explainable_2021}
Ning Wang, Mingxuan Chen, and K.~P. Subbalakshmi.
\newblock Explainable {CNN}-attention {Networks} ({C}-{Attention} {Network}) for {Automated} {Detection} of {Alzheimer}'s {Disease}, January 2021.
\newblock arXiv:2006.14135 [cs].

\bibitem{mikolov2013efficient}
Tomas Mikolov, Kai Chen, Greg Corrado, and Jeffrey Dean.
\newblock Efficient estimation of word representations in vector space.
\newblock {\em arXiv preprint arXiv:1301.3781}, 2013.

\bibitem{pennington2014glove}
Jeffrey Pennington, Richard Socher, and Christopher~D Manning.
\newblock Glove: Global vectors for word representation.
\newblock In {\em Proceedings of the 2014 conference on empirical methods in natural language processing (EMNLP)}, pages 1532--1543, 2014.

\bibitem{bojanowski2017enriching}
Piotr Bojanowski, Edouard Grave, Armand Joulin, and Tomas Mikolov.
\newblock Enriching word vectors with subword information.
\newblock {\em Transactions of the association for computational linguistics}, 5:135--146, 2017.

\bibitem{roshanzamir_transformer-based_2021}
Alireza Roshanzamir, Hamid Aghajan, and Mahdieh Soleymani~Baghshah.
\newblock Transformer-based deep neural network language models for {Alzheimer}’s disease risk assessment from targeted speech.
\newblock {\em BMC Medical Informatics and Decision Making}, 21(1):92, March 2021.

\bibitem{sun2020ernie}
Yu~Sun, Shuohuan Wang, Yukun Li, Shikun Feng, Hao Tian, Hua Wu, and Haifeng Wang.
\newblock Ernie 2.0: A continual pre-training framework for language understanding.
\newblock In {\em Proceedings of the AAAI conference on artificial intelligence}, volume~34, pages 8968--8975, 2020.

\bibitem{colla2022semantic}
Davide Colla, Matteo Delsanto, Marco Agosto, Benedetto Vitiello, and Daniele~P Radicioni.
\newblock Semantic coherence markers: The contribution of perplexity metrics.
\newblock {\em Artificial Intelligence in Medicine}, 134:102393, 2022.

\bibitem{saltz_dementia_2021}
Ploypaphat Saltz, Shih~Yin Lin, Sunny~Chieh Cheng, and Dong Si.
\newblock Dementia {Detection} using {Transformer}-{Based} {Deep} {Learning} and {Natural} {Language} {Processing} {Models}.
\newblock In {\em 2021 {IEEE} 9th {International} {Conference} on {Healthcare} {Informatics} ({ICHI})}, pages 509--510, August 2021.
\newblock ISSN: 2575-2634.

\bibitem{yang2019xlnet}
Zhilin Yang, Zihang Dai, Yiming Yang, Jaime Carbonell, Russ~R Salakhutdinov, and Quoc~V Le.
\newblock Xlnet: Generalized autoregressive pretraining for language understanding.
\newblock {\em Advances in neural information processing systems}, 32, 2019.

\bibitem{clark2020electra}
Kevin Clark, Minh-Thang Luong, Quoc~V Le, and Christopher~D Manning.
\newblock Electra: Pre-training text encoders as discriminators rather than generators.
\newblock {\em arXiv preprint arXiv:2003.10555}, 2020.

\bibitem{hong_novel_2019}
Sheng-Yi Hong, Li-Hung Yao, Wen-Ting Cheah, Wei-Der Chang, Li-Chen Fu, and Yu-Ling Chang.
\newblock A {Novel} {Screening} {System} for {Alzheimer}’s {Disease} {Based} on {Speech} {Transcripts} {Using} {Neural} {Network}.
\newblock In {\em 2019 {IEEE} {International} {Conference} on {Systems}, {Man} and {Cybernetics} ({SMC})}, pages 2440--2445, October 2019.
\newblock ISSN: 2577-1655.

\bibitem{mittal_multi-modal_2021}
Amish Mittal, Sourav Sahoo, Arnhav Datar, Juned Kadiwala, Hrithwik Shalu, and Jimson Mathew.
\newblock Multi-{Modal} {Detection} of {Alzheimer}'s {Disease} from {Speech} and {Text}, July 2021.
\newblock arXiv:2012.00096 [cs].

\bibitem{edwards_multiscale_2020}
Erik Edwards, Charles Dognin, Bajibabu Bollepalli, and Maneesh Singh.
\newblock Multiscale {System} for {Alzheimer}’s {Dementia} {Recognition} {Through} {Spontaneous} {Speech}.
\newblock In {\em Interspeech 2020}, pages 2197--2201. ISCA, October 2020.

\bibitem{balagopalan_comparing_2021}
Aparna Balagopalan, Benjamin Eyre, Jessica Robin, Frank Rudzicz, and Jekaterina Novikova.
\newblock Comparing {Pre}-trained and {Feature}-{Based} {Models} for {Prediction} of {Alzheimer}'s {Disease} {Based} on {Speech}.
\newblock {\em Frontiers in Aging Neuroscience}, 13, 2021.

\bibitem{rohanian_multi-modal_2020}
Morteza Rohanian, Julian Hough, and Matthew Purver.
\newblock Multi-{Modal} {Fusion} with {Gating} {Using} {Audio}, {Lexical} and {Disfluency} {Features} for {Alzheimer}’s {Dementia} {Recognition} from {Spontaneous} {Speech}.
\newblock In {\em Interspeech 2020}, pages 2187--2191. ISCA, October 2020.

\bibitem{ilias_multimodal_2022}
Loukas Ilias and Dimitris Askounis.
\newblock Multimodal {Deep} {Learning} {Models} for {Detecting} {Dementia} {From} {Speech} and {Transcripts}.
\newblock {\em Frontiers in Aging Neuroscience}, 14, 2022.

\bibitem{mahajan_acoustic_2021}
Pranav Mahajan and Veeky Baths.
\newblock Acoustic and {Language} {Based} {Deep} {Learning} {Approaches} for {Alzheimer}'s {Dementia} {Detection} {From} {Spontaneous} {Speech}.
\newblock {\em Frontiers in Aging Neuroscience}, 13, 2021.

\bibitem{rohanian_alzheimers_2021}
Morteza Rohanian, Julian Hough, and Matthew Purver.
\newblock Alzheimer's {Dementia} {Recognition} {Using} {Acoustic}, {Lexical}, {Disfluency} and {Speech} {Pause} {Features} {Robust} to {Noisy} {Inputs}, June 2021.
\newblock arXiv:2106.15684 [cs, eess].

\bibitem{gao2020survey}
Jing Gao, Peng Li, Zhikui Chen, and Jianing Zhang.
\newblock A survey on deep learning for multimodal data fusion.
\newblock {\em Neural Computation}, 32(5):829--864, 2020.

\bibitem{liu2019roberta}
Yinhan Liu, Myle Ott, Naman Goyal, Jingfei Du, Mandar Joshi, Danqi Chen, Omer Levy, Mike Lewis, Luke Zettlemoyer, and Veselin Stoyanov.
\newblock Roberta: A robustly optimized bert pretraining approach.
\newblock {\em arXiv preprint arXiv:1907.11692}, 2019.

\bibitem{sanh2019distilbert}
Victor Sanh, Lysandre Debut, Julien Chaumond, and Thomas Wolf.
\newblock Distilbert, a distilled version of bert: smaller, faster, cheaper and lighter.
\newblock {\em arXiv preprint arXiv:1910.01108}, 2019.

\bibitem{liu_transfer_2022}
Ning Liu, Kexue Luo, Zhenming Yuan, and Yan Chen.
\newblock A {Transfer} {Learning} {Method} for {Detecting} {Alzheimer}'s {Disease} {Based} on {Speech} and {Natural} {Language} {Processing}.
\newblock {\em Frontiers in Public Health}, 10, 2022.

\bibitem{jiang2022automated}
Zifan Jiang, Salman Seyedi, Rafi~U Haque, Alvince~L Pongos, Kayci~L Vickers, Cecelia~M Manzanares, James~J Lah, Allan~I Levey, and Gari~D Clifford.
\newblock Automated analysis of facial emotions in subjects with cognitive impairment.
\newblock {\em Plos one}, 17(1):e0262527, 2022.

\bibitem{alzahrani2021eye}
Fatimah Alzahrani, Bahman Mirheidari, Daniel Blackburn, Steve Maddock, and Heidi Christensen.
\newblock Eye blink rate based detection of cognitive impairment using in-the-wild data.
\newblock In {\em 2021 9th International Conference on Affective Computing and Intelligent Interaction (ACII)}, pages 1--8. IEEE, 2021.

\bibitem{tanaka2019detecting}
Hiroki Tanaka, Hiroyoshi Adachi, Hiroaki Kazui, Manabu Ikeda, Takashi Kudo, and Satoshi Nakamura.
\newblock Detecting dementia from face in human-agent interaction.
\newblock In {\em Adjunct of the 2019 International Conference on Multimodal Interaction}, pages 1--6, 2019.

\bibitem{lugaresi2019mediapipe}
Camillo Lugaresi, Jiuqiang Tang, Hadon Nash, Chris McClanahan, Esha Uboweja, Michael Hays, Fan Zhang, Chuo-Ling Chang, Ming~Guang Yong, Juhyun Lee, et~al.
\newblock Mediapipe: A framework for building perception pipelines.
\newblock {\em arXiv preprint arXiv:1906.08172}, 2019.

\bibitem{baltrusaitis2018openface}
Tadas Baltrusaitis, Amir Zadeh, Yao~Chong Lim, and Louis-Philippe Morency.
\newblock Openface 2.0: Facial behavior analysis toolkit.
\newblock In {\em 2018 13th IEEE international conference on automatic face \& gesture recognition (FG 2018)}, pages 59--66. IEEE, 2018.

\bibitem{fard2022ad}
Ali~Pourramezan Fard and Mohammad~H Mahoor.
\newblock Ad-corre: Adaptive correlation-based loss for facial expression recognition in the wild.
\newblock {\em IEEE Access}, 10:26756--26768, 2022.

\bibitem{fard2022sagittal}
Ali~Pourramezan Fard, Joe Ferrantelli, Anne-Lise Dupuis, and Mohammad~H Mahoor.
\newblock Sagittal cervical spine landmark point detection in x-ray using deep convolutional neural networks.
\newblock {\em IEEE Access}, 10:59413--59427, 2022.

\bibitem{fard2022facial}
Ali~Pourramezan Fard and Mohammad~H Mahoor.
\newblock Facial landmark points detection using knowledge distillation-based neural networks.
\newblock {\em Computer Vision and Image Understanding}, 215:103316, 2022.

\bibitem{fard2021asmnet}
Ali~Pourramezan Fard, Hojjat Abdollahi, and Mohammad Mahoor.
\newblock Asmnet: A lightweight deep neural network for face alignment and pose estimation.
\newblock In {\em Proceedings of the IEEE/CVF Conference on computer vision and pattern recognition}, pages 1521--1530, 2021.

\bibitem{lin2024data}
Emily Lin, Jian Sun, Hsingyu Chen, and Mohammad~H Mahoor.
\newblock Data quality matters: Suicide intention detection on social media posts using a roberta-cnn model.
\newblock {\em arXiv preprint arXiv:2402.02262}, 2024.

\bibitem{johnson2019survey}
Justin~M Johnson and Taghi~M Khoshgoftaar.
\newblock Survey on deep learning with class imbalance.
\newblock {\em Journal of Big Data}, 6(1):1--54, 2019.

\bibitem{mollahosseini2017affectnet}
Ali Mollahosseini, Behzad Hasani, and Mohammad~H Mahoor.
\newblock Affectnet: A database for facial expression, valence, and arousal computing in the wild.
\newblock {\em IEEE Transactions on Affective Computing}, 10(1):18--31, 2017.

\bibitem{f2023alyzer}
Ali~Pourramezan Fard, Mohammad~H Mahoor, Sarah~Ariel Lamer, and Timothy Sweeny.
\newblock Ganalyzer: Analysis and manipulation of gans latent space for controllable face synthesis.
\newblock {\em arXiv preprint arXiv:2302.00908}, 2023.

\bibitem{yampolskiy2019unexplainability}
Roman~V Yampolskiy.
\newblock Unexplainability and incomprehensibility of artificial intelligence.
\newblock {\em arXiv preprint arXiv:1907.03869}, 2019.

\bibitem{murdoch2019definitions}
W~James Murdoch, Chandan Singh, Karl Kumbier, Reza Abbasi-Asl, and Bin Yu.
\newblock Definitions, methods, and applications in interpretable machine learning.
\newblock {\em Proceedings of the National Academy of Sciences}, 116(44):22071--22080, 2019.

\bibitem{molnar2020interpretable}
Christoph Molnar, Giuseppe Casalicchio, and Bernd Bischl.
\newblock Interpretable machine learning--a brief history, state-of-the-art and challenges.
\newblock In {\em Joint European conference on machine learning and knowledge discovery in databases}, pages 417--431. Springer, 2020.

\bibitem{sun2023xnodr}
Jian Sun, Ali~Pourramezan Fard, and Mohammad~H Mahoor.
\newblock Xnodr and xnidr: Two accurate and fast fully connected layers for convolutional neural networks.
\newblock {\em Journal of Intelligent \& Robotic Systems}, 109(1):17, 2023.

\bibitem{khan2023secure}
Tahir~Abbas Khan, Areej Fatima, Tariq Shahzad, Khalid Alissa, Taher~M Ghazal, Mahmoud~M Al-Sakhnini, Sagheer Abbas, Muhammad~Adnan Khan, Arfan Ahmed, et~al.
\newblock Secure iomt for disease prediction empowered with transfer learning in healthcare 5.0, the concept and case study.
\newblock {\em IEEE Access}, 11:39418--39430, 2023.

\bibitem{zhu2023transfer}
Zhuangdi Zhu, Kaixiang Lin, Anil~K Jain, and Jiayu Zhou.
\newblock Transfer learning in deep reinforcement learning: A survey.
\newblock {\em IEEE Transactions on Pattern Analysis and Machine Intelligence}, 2023.

\bibitem{kheddar2023deep}
Hamza Kheddar, Yassine Himeur, Somaya Al-Maadeed, Abbes Amira, and Faycal Bensaali.
\newblock Deep transfer learning for automatic speech recognition: Towards better generalization.
\newblock {\em Knowledge-Based Systems}, 277:110851, 2023.

\bibitem{you2022ranking}
Kaichao You, Yong Liu, Ziyang Zhang, Jianmin Wang, Michael~I Jordan, and Mingsheng Long.
\newblock Ranking and tuning pre-trained models: A new paradigm for exploiting model hubs.
\newblock {\em Journal of Machine Learning Research}, 23(209):1--47, 2022.

\bibitem{deng2009imagenet}
Jia Deng, Wei Dong, Richard Socher, Li-Jia Li, Kai Li, and Li~Fei-Fei.
\newblock Imagenet: A large-scale hierarchical image database.
\newblock In {\em 2009 IEEE conference on computer vision and pattern recognition}, pages 248--255. Ieee, 2009.

\bibitem{wen2021rethinking}
Yang Wen, Leiting Chen, Yu~Deng, and Chuan Zhou.
\newblock Rethinking pre-training on medical imaging.
\newblock {\em Journal of Visual Communication and Image Representation}, 78:103145, 2021.

\bibitem{samann2022freesurfer}
Philipp~G S{\"a}mann, Juan~Eugenio Iglesias, Boris Gutman, Dominik Grotegerd, Ramona Leenings, Claas Flint, Udo Dannlowski, Emily~K Clarke-Rubright, Rajendra~A Morey, Theo~GM van Erp, et~al.
\newblock Freesurfer-based segmentation of hippocampal subfields: A review of methods and applications, with a novel quality control procedure for enigma studies and other collaborative efforts.
\newblock {\em Human brain mapping}, 43(1):207--233, 2022.

\bibitem{holzinger2017we}
Andreas Holzinger, Chris Biemann, Constantinos~S Pattichis, and Douglas~B Kell.
\newblock What do we need to build explainable ai systems for the medical domain?
\newblock {\em arXiv preprint arXiv:1712.09923}, 2017.

\bibitem{albahri2023systematic}
Ahmed~Shihab Albahri, Ali~M Duhaim, Mohammed~A Fadhel, Alhamzah Alnoor, Noor~S Baqer, Laith Alzubaidi, Osamah~Shihab Albahri, Abdullah~Hussein Alamoodi, Jinshuai Bai, Asma Salhi, et~al.
\newblock A systematic review of trustworthy and explainable artificial intelligence in healthcare: Assessment of quality, bias risk, and data fusion.
\newblock {\em Information Fusion}, 2023.

\bibitem{zhang2021survey}
Chen Zhang, Yu~Xie, Hang Bai, Bin Yu, Weihong Li, and Yuan Gao.
\newblock A survey on federated learning.
\newblock {\em Knowledge-Based Systems}, 216:106775, 2021.

\end{thebibliography}



\end{document}